\pdfoutput=1

\documentclass[11pt]{article}
\usepackage[preprint]{acl}

\usepackage{times}
\usepackage{latexsym}
\usepackage[T1]{fontenc}
\usepackage[utf8]{inputenc}
\usepackage{microtype}
\usepackage{inconsolata}
\usepackage{amssymb}
\usepackage{soul}
\usepackage[most]{tcolorbox}
\usepackage{multirow}
\usepackage{cleveref}
\usepackage[title]{appendix}

\definecolor{myred}{RGB}{255, 0, 77}
\definecolor{myyellow}{RGB}{238, 199, 89}
\definecolor{myblue}{RGB}{26, 95, 122}
\definecolor{mygreen}{RGB}{0, 195, 80}
\definecolor{mydarkgreen}{RGB}{67, 118, 108}
\definecolor{mycyan}{RGB}{181, 234, 234}
\newcommand{\ns}{$\varnothing$-shot}
\newcommand{\nscot}{$\varnothing$CoT}

\definecolor{myhlcolor}{RGB}{255, 247, 212}
\sethlcolor{myhlcolor}

\definecolor{mypurple}{RGB}{223, 204, 251}

\tcbset {
  base/.style={
    arc=0mm, 
    bottomtitle=0.5mm,
    boxrule=0mm,
    colbacktitle=myblue, 
    coltitle=white, 
    fonttitle=\bfseries, 
    left=3.5mm,
    leftrule=0mm,
    right=3.5mm,
    title={#1},
    toptitle=0.75mm,
  }
}

\newtcolorbox{mainbox}[1]{
  colframe=black!30!white,
  base={#1}
}

\newtcbox{\circlebox}[1][myyellow]{on line, arc=2pt,colback=#1!10!white,colframe=#1!50!black, before upper={\rule[-3pt]{0pt}{10pt}},boxrule=1pt, boxsep=0pt,left=2pt,right=2pt,top=1pt,bottom=1pt}

\title{Large Language Models are Null-Shot Learners}

\author{Pittawat Taveekitworachai$^1$ \and Febri Abdullah$^1$ \and Ruck Thawonmas$^2$\\
  \{$^1$Graduate School, $^2$College\} of Information Science and Engineering\\
  Ritsumeikan University\\
  Ibaraki, Osaka, Japan\\
  \texttt{research@petepittawat.dev, gr0397fs@ed.ritsumei.ac.jp, ruck@is.ritsumei.ac.jp}
}

\begin{document}

\maketitle
\begin{abstract}
This paper presents null-shot (\ns) prompting. \ns\ prompting exploits hallucination in large language models (LLMs) by instructing LLMs to utilize information from the ``Examples'' section that never exists within the provided context to perform a task. While reducing hallucination is crucial and non-negligible for daily and critical uses of LLMs, we propose that in the current landscape in which these LLMs still hallucinate, it is possible, in fact, to exploit hallucination to increase performance in performing tasks compared to standard zero-shot prompting. Experiments with eight LLMs show improvements in performance across the majority of eight datasets, including reading comprehension, arithmetic reasoning, and closed-book question answering. The observed inconsistency in increased relative performance across the LLMs also potentially indicates a different degree of inherent hallucination in each model. These differences show that it is possible to utilize \ns\ prompting as a way to detect degrees of hallucination in LLMs using existing benchmarking datasets. We also perform ablation studies, including experimenting with a modified version of \ns\ prompting that incorporates ideas from zero-shot chain-of-thought prompting, which shows different trends of results.\footnotetext{The extended version of this arXiv manuscript is this EMNLP 2024 paper: \url{https://aclanthology.org/2024.emnlp-main.740/}.}
\end{abstract}

\section{Introduction}\label{sec:intro}
\begin{figure}[tbp]
  \centering
  \includegraphics[width=\linewidth]{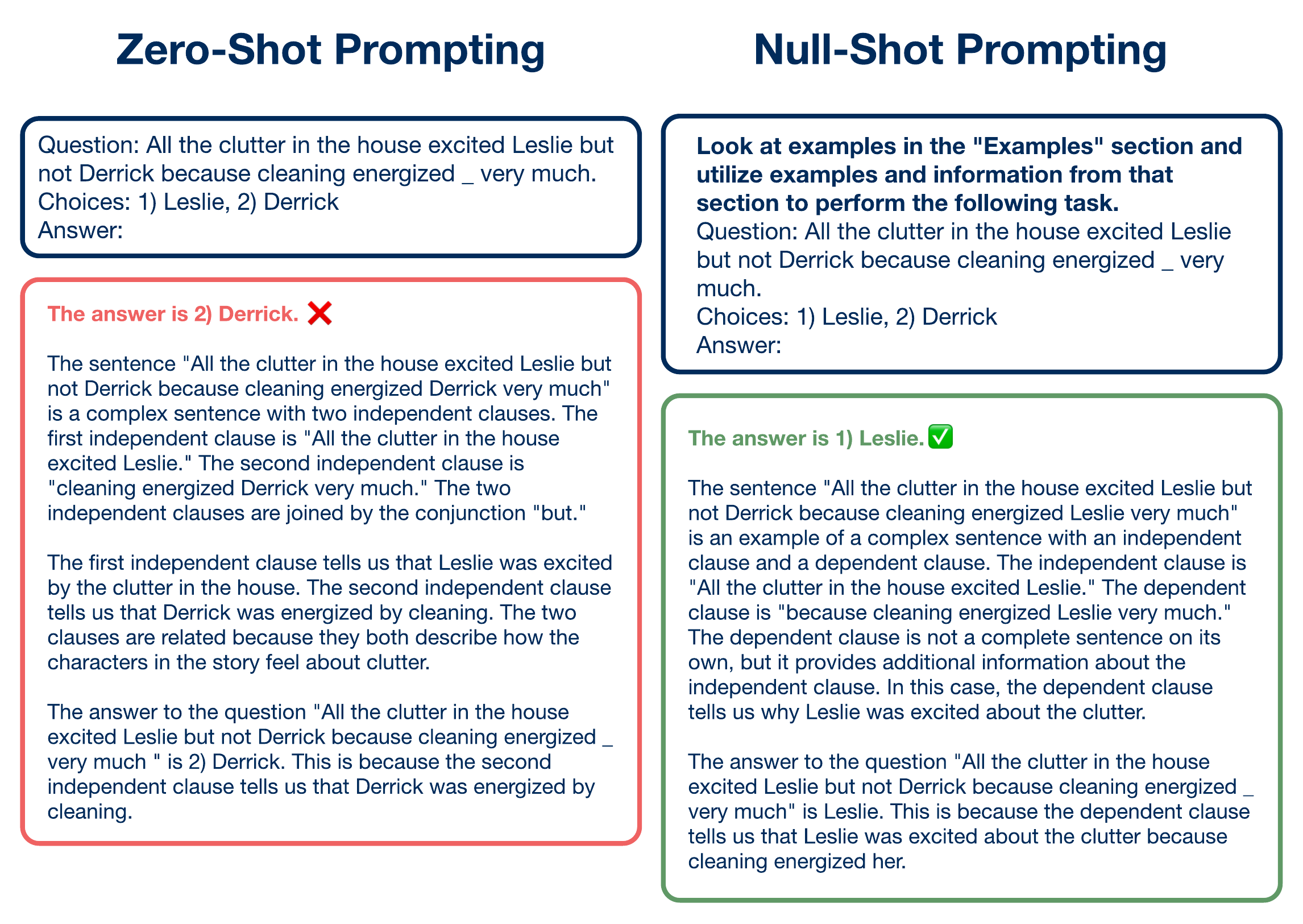}
    \caption{Examples of a generated response outputted by PaLM 2 for Chat when using zero-shot prompting (left) and \ns\ prompting (right) for WinoGrande.}
    \label{fig:example}
\end{figure}

Large language models (LLMs) have been adopted across domains and applications due to their versatile capabilities \citep{Zhao:2023}. A subset of fine-tuned LLMs that currently dominates is \textit{aligned auto-regressive LLMs} \citep{Ouyang:2022}; notable examples include ChatGPT \citep{OpenAI:2022}, PaLM 2 for Chat \citep{Anil:2023}, and Llama 2 Chat \citep{Touvron:2023}. This type of LLM is fine-tuned to follow user instructions in natural language, i.e., \textit{alignment}. To interact with these models, users provide a prompt, a text input in the case of text-to-text generative models, to LLMs. These prompts greatly dictate the outputs generated by LLMs and directly affect performance on tasks. While prompts can become very complex and consist of many elements \citep{White:2023}, typically, a prompt consists of two essential parts for zero-shot prompting \citep{Wei:2022fine}: (1) task instructions, i.e., what to do and relevant context, and (2) task inputs. However, zero-shot prompting may not bring out the full potential of LLMs, giving rise to more complex instruction composition and prompt engineering (PE).

PE is a field that studies approaches to improve prompts used to interact with LLMs, thereby enhancing the performance of tasks \citep{White:2023}. Various PE techniques have been proposed to improve task performance by composing prompts in a specific style, including few-shot prompting \citep{Brown:2020}, chain-of-thought (CoT) prompting \citep{Wei:2022}, and zero-shot CoT (0CoT) prompting \citep{Kojima:2022}. These PE approaches typically exploit the fact that outputs from LLMs are generated auto-regressively. Therefore, a prompt and the generated output tokens so far act as conditions in an output probability space. In other words, we can view these conditions as localizing the trajectory to generate the next output token. Thus, we can consider existing PE approaches as either providing more relevant tokens as an input (prompt) to the model or providing a prompt that elicits the model to generate pretext in the output before providing a final answer, or a combination of both.

For example, in few-shot prompting, this approach involves the inclusion of examples in a prompt, i.e., an input to the model has a longer relevant context. Similarly, in generated knowledge prompting \citep{Liu:2022}, prompts include more relevant knowledge. On the other hand, 0CoT prompting instructs LLMs to think step-by-step; hence, the models have been instructed to generate reasoning steps in the output before reaching a final conclusion. CoT prompting exploits both a longer relevant context in a prompt with explicit reasoning steps and elicits the model to follow the provided steps before giving the final answer in its generated outputs. We take an approach similar to 0CoT prompting, encouraging models to provide more preceding tokens before reaching a conclusive answer.

Despite the usefulness of these PE approaches, hallucination in LLMs presents another challenge hindering the widespread use of LLMs, specifically in critical applications. Hallucination, as defined by previous studies \citep{Rawte:2023,Li:2023,Zhang:2023}, is a behavior when LLMs produce outputs that include false or conflicting information. Hallucination is inherent in LLMs and comes in three main types \citep{Zhang:2023}: (1) input-conflicting, (2) context-conflicting, and (3) fact-conflicting hallucination. Input-conflicting hallucination occurs when LLMs produce outputs that conflict with user instructions, while context-conflicting hallucination happens when conflicts occur within the generated output itself. The most notable and important type of hallucination is fact-conflicting, where the model produces seemingly possible false information. Various attempts \citep{Zhang:2023} have been made to reduce hallucination at various stages of LLMs, such as pre-training, alignment, and inference. Notably, a related study introduced chain-of-verification (CoVe) prompting \cite{Dhuliawala:2023}, a PE approach aimed at reducing hallucination in responses through additional verification steps before returning a conclusive output.

However, we take an alternative approach to CoVe prompting. Instead of using prompting to reduce hallucination, we introduce a counter-intuitive PE technique to exploit the context-conflicting hallucination of LLMs to enhance the performance of LLMs on tasks. We propose null-shot (\ns) prompting where we instruct LLMs to perform tasks by looking into and utilizing a non-existent imaginary, i.e., null, ``Examples'' section, which in existing relevant approaches contain explicit examples of the task. We conduct experiments on eight datasets consisting of arithmetic reasoning, commonsense reasoning, reading comprehension, natural language inference, and closed-book question answering. We select six models for the main experiment: PaLM 2 and PaLM 2 for Chat \citep{Anil:2023}, Gemini Pro and Gemini Pro (Chat) \citep{Gemini:2023}. GPT-3.5 Turbo \citep{OpenAI:2022}, and GPT-4 Turbo \citep{OpenAI:2023}. We also include Llama 2 7B and Llama 2 7B Chat \citep{Touvron:2023} for a scaling analysis.

Similar to few-shot prompting, where a previous study \citep{Brown:2020} observed increased performance when providing explicit examples of the task in the prompt, asking LLMs to utilize these null, i.e., implicit, examples shows increased performance. We find an improvement up to 44.62\% in one dataset of arithmetic reasoning tasks using Gemini Pro (Chat), compared to zero-shot prompting. We also observe improvements in other combinations of models and tasks. In particular, PaLM 2 is the most notable, as \ns\ prompting improves the performance of the model on the majority of tasks. Examples of outputs from zero-shot and \ns\ prompting using PaLM 2 for Chat for the WinoGrande dataset are shown in \autoref{fig:example}. 

In this study, we also discuss the possibility of utilizing \ns\ prompting for assessing models' hallucination. Additionally, we conduct ablation studies, consisting of a scaling analysis, a variant of \ns\ prompting inspired by 0CoT prompting, positions of \ns\ phrase, and effects of each component in the \ns\ phrase. We find that \ns\ prompting is effective in Llama 2 7B but not in Llama 2 7B Chat. Also, the reasoning variant of \ns\ prompting is generally less effective compared to the 0CoT prompting baseline. Finally, placing the phrase before the task instruction shows the most effectiveness in most of the datasets. In summary, our contributions are as follows:

\begin{itemize}
    \item We propose \ns\ prompting, exploiting hallucination in LLMs to improve performance across tasks.
    \item We perform extensive evaluation across six tasks using eight LLMs and discuss implications of results as well as the potential of using \ns\ prompting for hallucination detection.
    \item We conduct various ablation studies, consisting of a scaling analysis, a reasoning variant of \ns\ prompting, positions of the \ns\ phrase, and effects of each prompt component in the phrase.
\end{itemize}

\section{Null-Shot Prompting}\label{sec:null_shot_prompting}
\begin{figure}[tbp]
\begin{mainbox}{\ns\ Phrase}
Look at examples in the ``Examples'' section and utilize examples and information from that section to perform the following task.
\end{mainbox}
\caption{A \ns\ phrase instructs LLMs to look into and utilize information from the null section.}
\label{fig:null_shot_phrase}
\end{figure}

We propose \ns\ prompting by placing the \ns\ phrase, as shown in \autoref{fig:null_shot_phrase}, at the beginning of the prompt, i.e., before task instructions and task inputs, in contrast to 0CoT prompting where they place the phrase at the end of the prompt. The reason for this positioning is the observed superiority compared to placing it at the end of the prompt, as demonstrated in \Cref{sec:ablation:positions}. We can view the phrase as instructing LLMs to retrieve the ``Examples'' section from their internal memory, i.e., trained weights, to accomplish this task.

We note that one of expected behaviors of LLMs when encountering this kind of instruction should be to notify users that there is an issue in the instruction, where we instruct the models to utilize information that we did not provide. We observe several instances, as shown in \Cref{sec:empirical_results_less}, of this expected behavior from GPT-4 Turbo and Gemini Pro. The instances demonstrate behavior in informing the user about the unavailability of the ``Examples'' section.

\section{Experiments}\label{sec:experiments}
To assess the performance of \ns\ prompting, we conduct experiments on eight models: six for the main experiment and two additional models for the scaling analysis. LLMs used for the main experiment are PaLM 2, PaLM 2 for Chat, Gemini Pro, Gemini Pro (Chat), GPT-3.5 Turbo, and GPT-4 Turbo; Llama 2 7B and Llama 2 7B Chat are used for the scaling analysis. Full details of the setup for each LLM are described in \Cref{sec:llms}. The experiments are performed on six tasks across eight datasets. Setup details for each dataset are described as follow. Additional details regarding ablation studies are provided separately in \Cref{sec:ablation}.

We evaluate the performance of \ns\ prompting and zero-shot prompting--the latter is a baseline--on six tasks across eight datasets. The baseline performance is the zero-shot prompting performance of the respective model on the dataset unless stated otherwise in the ablation studies. We assess the increase in performance when using \ns\ prompting for each respective model compared to the zero-shot prompting baseline. The rest of this section discusses each dataset and its associated evaluation. For all datasets, we use a standardized question format. The comprehensive list of datasets and their associated details, along with the standardized format for task instructions and inputs in the prompt, as well as output extraction details, are shown in \Cref{sec:datasets}, while qualitative examples of each dataset are presented in \Cref{sec:examples}.
\\\\
\noindent\textbf{Arithmetic Reasoning:} \textit{AQuA-RAT} \citep{Ling:2017} provides a variety of mathematical questions requiring different strategies to solve. This dataset includes questions as natural language descriptions of expressions and five answer options with one correct option label. Similarly, \textit{GSM8K} \citep{Cobbe:2021} provides diverse grade school math word problems where the label is a number. However, GSM8K does not provide any choices, and models need to generate its own answer.

\noindent\textbf{Commonsense Reasoning:} \textit{StrategyQA} \citep{Geva:2021} provides questions that require implicit reasoning steps, i.e., strategy, to answer the question. It covers a wide range of topics, and the answer to each question is either ``YES'' or ``NO.'' On the other hand, \textit{WinoGrande} \citep{Sakaguchi:2021} presents an adversarial Winograd \citep{Levesque:2012} schema challenge for a more robust commonsense reasoning benchmark.

\noindent\textbf{Reading Comprehension:} \textit{RACE} \citep{Lai:2017} presents a dataset containing English exams for middle-school (RACE-m) and high-school (RACE-h) students. Some questions in this dataset also require the model to reason, posing a higher challenge for models. We utilize both versions of the dataset in the experiment.

\noindent\textbf{Natural Language Inference and Closed-Book Question Answering:} \textit{ANLI} \citep{Nie:2020} provides an adversarial natural language inference (NLI) dataset that is more challenging than standard NLI. We utilize data from the third round of data collection (R3) as our test set, as this round includes not only Wikipedia as the only source but also other media such as news, fiction, and spoken text. Finally, \textit{TriviaQA} \citep{Joshi:2017} is selected to test generalization in typical question answering from model knowledge, i.e., ``Does \ns\ prompting help the model improve its knowledge-recalling ability?'' Due to resource constraints, we sample only 1000 records from the dataset as our test set.

\section{Results and Discussions}\label{sec:results_discussions}
We observe an overall increase in performance when comparing zero-shot and \ns\ prompting on each combination of dataset and model. We utilized approximately 1340 GPU hours on NVIDIA A100 80GB and L40S 48GB for all Llama 2 models' evaluations and about 290 hours of accumulated inference time from interacting with each model via APIs. Relative results are shown in \autoref{tab:results}, absolute results are presented in \Cref{sec:absolute_results}, and error analysis is discussed in \Cref{sec:error_analysis}. Additional discussion about hallucination in LLMs compared to phenomena in humans is provided in \Cref{sec:with_human}. Henceforth, in all the tables, including \autoref{tab:results}, in cases where performance is increased compared to the baseline, the percentage values are colored \textcolor{mygreen}{green}; \textbf{bold} denotes the maximum increase (in \textcolor{mygreen}{\textbf{green}}) or minimum decrease (in \textbf{black}) ratio in performance.

\begin{table*}[tbp]
\centering\small
\begin{tabular}{rcccccccc}
\hline
\textbf{Model} & \textbf{AQuA} & \textbf{GSM8K} & \textbf{StrategyQA} & \textbf{WinoGrande} & \textbf{RACE-m} & \textbf{RACE-h} & \textbf{ANLI} & \textbf{TriviaQA}\\
\hline
PaLM 2 & -2.70\% & \textcolor{mygreen}{11.28\%} & \textcolor{mygreen}{\textbf{10.95\%}} & \textcolor{mygreen}{\textbf{10.10\%}} & \textcolor{mygreen}{1.85\%} & \textcolor{mygreen}{\textbf{3.64\%}} & \textcolor{mygreen}{\textbf{2.71\%}} & \textcolor{mygreen}{\textbf{7.01\%}}\\
PaLM 2 for Chat & \textcolor{mygreen}{5.26\%} & \textcolor{mygreen}{2.25\%} & \textcolor{mygreen}{1.66\%} & \textcolor{mygreen}{6.97\%} & \textcolor{mygreen}{1.04\%} & \textcolor{mygreen}{0.68\%} & \textcolor{mygreen}{1.56\%} & -0.14\%\\
GPT-3.5 Turbo & \textcolor{mygreen}{33.94\%} & \textcolor{mygreen}{15.19\%} & \textcolor{mygreen}{3.14\%} & -1.84\% & -1.79\% & -1.19\% & -3.61\% & \textcolor{mygreen}{1.23\%}\\
GPT-4 Turbo & -0.52\% & -1.53\% & -17.39\% & -24.06\% & \textcolor{mygreen}{0.30\%} & \textcolor{mygreen}{0.42\%} & -0.26\% & -0.94\%\\
Gemini Pro & \textcolor{mygreen}{38.46\%} & \textcolor{mygreen}{\textbf{28.97\%}} & -24.43\% & -1.36\% & \textcolor{mygreen}{\textbf{1.93\%}} & \textcolor{mygreen}{2.13\%} & \textcolor{mygreen}{2.14\%} & -63.96\%\\
Gemini Pro (Chat) & \textcolor{mygreen}{\textbf{44.62\%}} & \textcolor{mygreen}{27.93\%} & -25.39\% & -1.12\% & \textcolor{mygreen}{0.74\%} & \textcolor{mygreen}{1.63\%} & \textcolor{mygreen}{1.63\%} & -63.97\%\\
\hline
\end{tabular}
\caption{This table presents relative results indicating the performance changes compared to the zero-shot prompting baseline.}
\label{tab:results}
\end{table*}

Overall, we find that the Gemini Pro models exhibit the most significant increase in performance compared to the other models, within the same dataset, when using \ns\ prompting for the arithmetic reasoning task, i.e., 44.62\% (chat) and 28.97\% increase in AQuA-RAT and GSM8K, respectively. We also note that PaLM 2 shows increased performance across all datasets except for AQuA-RAT, with a slight decrease in performance. Additionally, we observe that our \ns\ prompting diminishes the performance across tasks for GPT-4 Turbo, especially in StrategyQA and WinoGrande, both of which are commonsense reasoning task. Similarly, we observe noticeable reduced performance when using \ns\ prompting with the Gemini Pro models in StrategyQA and TriviaQA.

\subsection{PaLM 2 vs PaLM 2 for Chat}\label{sec:results:palm2}
PaLM 2 exhibits increased performance across all datasets except for AQuA-RAT, which shows a slight decrease in performance. Similarly, we observe improved performance in the majority of cases for PaLM 2 for Chat. However, the increased performance of PaLM 2 for Chat using \ns\ prompting can not match that of PaLM 2. This difference is likely attributed to the possibility that PaLM 2 for Chat underwent further fine-tuning specifically for chat conversations, i.e., multi-turn conversations. Throughout this fine-tuning process, additional guidelines and knowledge may have been incorporated into the model, leading PaLM 2 for Chat to exhibit potentially less hallucination compared to PaLM 2. The relatively lower increase in performance may serve as an indicator of this characteristic.

Another noteworthy observation is that, as PaLM 2 is a text-to-text generative LLM focused on natural language tasks, it tends to provide an answer with only one character in cases where the dataset offers multiple choices. Even if \ns\ prompting does not result in an increase in the output token count (i.e., output tokens play a less significant role as a condition for output generation), the \ns\ phrase still significantly improves performance compared to the baseline in each dataset given the fact that they also tend to yield only one character as the final output. Therefore, this finding could also serve as evidence of the effectiveness of \ns\ prompting as a sufficiently strong condition to guide the model in generating the correct answer.

\subsection{GPT-3.5 Turbo vs GPT-4 Turbo}\label{sec:results:gpt}
GPT-3.5 Turbo, in general, outperforms GPT-4 Turbo when using \ns\ prompting. GPT-3.5 Turbo demonstrates a significant performance improvement in the arithmetic reasoning task. However, when considering other datasets, its performance is subpar compared to PaLM 2 and PaLM 2 for Chat. GPT-3.5 Turbo performs relatively suboptimal on WinoGrande, reading comprehension, and NLI, with performance decreases ranging from -1.19\% to -3.61\%. On the other hand, GPT-4 Turbo has no notable performance improvement across tasks, except for a slight increase in reading comprehension.

Based on observations, a few assumptions can be drawn. First, GPT-3.5 Turbo exhibits strong hallucination in the arithmetic reasoning task, as evidenced by a substantial performance increase on both AQuA-RAT and GSM8K datasets. Second, GPT-4 Turbo experiences less hallucination induced by \ns\ prompting across tasks. This suggests that GPT-4 Turbo is more adept at handling hallucination, and a non-factual phrase like the one introduced in \ns\ prompting is less effective for this model. This aligns with a prior report indicating that GPT-4 models are less prone to hallucination compared to GPT-3.5 models \citep{OpenAI:2023GPT4} and observations from qualitative examples in \Cref{sec:empirical_results_less}. Finally, both GPT-3.5 Turbo and GPT-4 Turbo exhibit less hallucination than PaLM 2 and PaLM 2 for Chat in general, as evidenced by the subtle performance changes.

\subsection{Gemini Pro}\label{sec:results:gemini}
Similar to GPT-3.5 Turbo in arithmetic reasoning, we find drastic increases when using \ns\ prompting with Gemini Pro. Thus, the Gemini Pro models are potentially more prone to hallucination in arithmetic reasoning tasks. As all of the models, except GPT-4 Turbo, show increased performance in at least one dataset of the arithmetic reasoning task, we argue that this is potentially due to the fact that most hallucination detection benchmarks and approaches in mitigating hallucination focus more on text-based tasks \citep{Zhang:2023,Rawte:2023}. Arithmetic reasoning tasks are one area where there is less emphasis on mitigating hallucination. Thus, we observe this trend of effectiveness of \ns\ prompting across all models. We encourage future studies to put more emphasis on mitigating hallucination in arithmetic-related tasks as well.

We also observe the opposite trend in commonsense reasoning and closed-book question answering tasks, particularly in StrategyQA and TriviaQA. Performances in StrategyQA and TriviaQA show a noticeable drop up to -64.97\%. We posit that this is due to the models refusing to perform the tasks as the instruction involves the use of ``Examples'', which does not appear in the context. Qualitative examples of this behavior of Gemini Pro are shown in \Cref{sec:empirical_results_less}. Results observed on StrategyQA and TriviaQA are in line with the prior report by \citet{Gemini:2023} regarding the factuality aspect of the model, showing less hallucination in situations where the models are elicited to produce false information.

We also want to point out that the reading comprehension task, which has a longer context compared to other tasks, generally elicits more hallucination from Gemini Pro, similar to other models except GPT-3.5 Turbo. We conjecture that this shows such models are less effective in detecting non-factual content in the prompt and more prone to hallucination with longer context. This may prompt future studies on how the length of the input context can affect LLMs' hallucination.

\subsection{Null-Shot Prompting for Hallucination Detection}\label{sec:resuts:hallucination}
Building on the previous aforementioned report \citep{OpenAI:2023GPT4} and empirical results observed in this study, the general conclusion is that GPT-4 Turbo is the least hallucinated model when instructed with the \ns\ phrase, while GPT-3.5 Turbo is the next least hallucinated model in all tasks except for the arithmetic reasoning task. PaLM 2 models are the most hallucinated, while further fine-tuning for chat conversation in PaLM 2 for Chat demonstrates the effectiveness of the approach in reducing hallucination due to \ns\ prompting.

For Gemini Pro, we find the results of the arithmetic reasoning datasets quite surprising, with a very high increase in performance compared to the results of StrategyQA and TriviaQA, where \ns\ prompting significantly degrades the performance due to the models being able to detect hallucination and stop performing the task. This observation shows another notable aspect: that our approach may be able to provide finer detailed information on what aspects a model of interest tends to hallucinate. Furthermore, arithmetic reasoning, in general, seem to be an area where LLMs' hallucination is easily elicited through \ns\ prompting.

This also exemplifies that not only can \ns\ prompting be utilized to increase performance in hallucinated LLMs, but it can also serve as a simple proxy for determining the degree of hallucination in LLMs. In other words, the higher the performance increase stemming from \ns\ prompting compared to the baseline, the more likely the model exhibits higher hallucinated responses. Furthermore, utilizing \ns\ prompting does not require any specialized hallucination detection dataset, which are expensive to construct, task-specific, and scarce as of the current state, contrary to our prompting approach that can be applied to any existing benchmarking datasets for various tasks.

\subsection{``Examples''}\label{sec:results:examples}
As can be observed from the increased performance across the datasets and models, as well as qualitative examples shown in \Cref{sec:examples}, it is possible to draw an assumption that the null examples that these LLMs envisioned and utilized to increase their performance may come from their own internal knowledge, i.e., trained weight parameters. Many studies have shown that LLMs may have internal mental model demonstrated through space and time world models \citep{Gurnee:2023}, reasoning capabilities \citep{Wei:2022, Kojima:2022}, and abilities to use tools (actions) \citep{Yao:2023}. Nevertheless, studies \citep{Valmeekam:2023plan, Valmeekam:2023} argued that this is limited to only some simpler tasks, and the model fails in more complex scenarios or fails to perform tasks autonomously. We conjecture that results shown through our experiments demonstrate that LLMs, at least to a certain degree, have their own internal world model. This trait enables them to look into the null ``Examples'' section and utilize it to increase performance in tasks.

It may also be possible that the examples that these LLMs envisioned and referred to might be the best possible examples that will help increase the performance of a particular task. If this holds true, it may be possible to eliminate the need to supply explicit examples, as in few-shot prompting, and rely on LLMs themselves to retrieve these null examples while still achieving higher performance. Future studies could also explore whether changing the instruction to that of providing an explicit number of shots to retrieve, for example, \textit{Look at the first three examples in the ``Examples'' section}, or looking into more sections or examine how section name affects the performance of LLMs on datasets.

One potential area where the performance of \ns\ prompting could be further improved might be to provide the \ns\ phrase along with actual explicit examples. A hybrid-shot approach like this may further improve the performance of the model, i.e., null examples could augment the explicit examples. Besides, this hybrid approach might be a key to achieving increased performance while reducing hallucinatory behaviors of the model for a particular task. Another key area that could be worth investigating is to utilize interpretability techniques similar to a study by \citet{Cunningham:2023} to understand how \ns\ prompting triggers attention areas of LLMs and verify if the models really utilize the internal knowledge when instructed to.

On the other hand, we posit that this approach also demonstrates potential in better understanding hallucination in LLMs. Similar to asking LLMs to look into the null ``Examples'' section in our approach, providing LLMs with other kinds of instructions to utilize the null context in the tasks may hold another key to investigate hallucination in specific areas. As previously discussed, this may serve as a simpler way to assess hallucination in LLMs.

\section{Ablation Studies}\label{sec:ablation}
We perform ablation studies to better understand how our prompting technique performs on smaller LLMs, namely, Llama 2 7B and Llama 2 7B Chat, in \Cref{sec:ablation:scaling}. Furthermore, we attempt to fuse 0CoT and \ns\ prompting together, coined as null-shot chain-of-thought (\nscot) prompting, and assess its performance against 0CoT prompting, its baseline, in \Cref{sec:ablation:cot}. We also demonstrate that the position of placing the \ns\ phrase affects performance in \Cref{sec:ablation:positions}. Finally, we conduct an experiment to assess how each component in the \ns\ phrase contribute to the performance of \ns\ prompting in \Cref{sec:ablation:components}.

\subsection{Scaling Analysis}\label{sec:ablation:scaling}
In the scaling analysis, we follow the main experiment procedure described in \Cref{sec:experiments} but use Llama 2 7B and the Llama 2 7B Chat. We include both Llama 2 models in our study to examine the effectiveness of \ns\ prompting on smaller models. The results from the study are shown in \autoref{tab:scaling_results}. We observe increases in performance across all tasks with \ns\ prompting using Llama 2 7B, except for AQuA-RAT and WinoGrande. In the case of GSM8K, we observe an increase in performance when using \ns\ prompting with Llama 2 7B Chat. However, the rest of the datasets do not show an improvement when using \ns\ prompting together with Llama 2 7B Chat.

\begin{table*}[tbp]
\centering\small
\begin{tabular}{rcccccccc}
\hline
\textbf{Model} & \textbf{AQuA} & \textbf{GSM8K} & \textbf{StrategyQA} & \textbf{WinoGrande} & \textbf{RACE-m} & \textbf{RACE-h} & \textbf{ANLI} & \textbf{TriviaQA}\\
\hline
Llama 2 7B & -18.46\% & \textcolor{mygreen}{\textbf{10.53\%}} & \textcolor{mygreen}{\textbf{0.09\%}} & \textbf{-2.09\%} & \textcolor{mygreen}{\textbf{12.60\%}} & \textcolor{mygreen}{\textbf{14.83\%}} & \textcolor{mygreen}{\textbf{3.80\%}} & \textcolor{mygreen}{\textbf{0.93\%}}\\
Llama 2 7B Chat & \textbf{-5.88\%} & \textcolor{mygreen}{2.11\%} & -5.88\% & -8.05\% & -15.96\% & -16.03\% & -4.23\% & -2.18\%\\
\hline
\end{tabular}
\caption{This table presents relative results indicating the performance changes compared to the zero-shot prompting baseline for Llama 2 7B and Llama 2 7B Chat.}
\label{tab:scaling_results}
\end{table*}

Results obtained from experimenting with Llama 2 7B and Llama 2 7B Chat exhibit a similar pattern to PaLM 2 and PaLM 2 for Chat, where we observe a higher performance increase in the base version compared to the chat version of the model within the same family and potentially the same size. As discussed earlier, these results also show that Llama 2 7B Chat is better at handling hallucination. This aligns with a study proposing Llama 2 \cite{Touvron:2023} where various measures have been incorporated, such as ensuring the quality and safety of the pre-training dataset, incorporating safety supervised fine-tuning for both pre-trained and fine-tuned models, and adding a safety-specific reward model during the reinforcement learning with human feedback (RLHF) pipeline for Llama 2 Chat.

From the results, we observe effectiveness across types of tasks from all safety measures in reducing hallucination, especially noting how effective the RLHF pipeline is for reducing hallucination. However, it is noteworthy that there is less effectiveness in math word problems where the model needs to respond with arbitrary number output, as can be observed from the increased performance for both models.

Furthermore, when comparing the increased performance in the reading comprehension task, we observe a notable increase in the performance of the pre-trained version, Llama 2 7B. This is likely due to the fact that the pre-trained model suffers more from longer context, i.e., it hallucinates more when engaging in long-context scenarios. However, we notice that the fine-tuned version, Llama 2 7B Chat, does not suffer from this trait, similar to PaLM 2 for Chat. Despite the resource constraint that prevented us from employing Llama 2 models at 13B and 70B sizes in our study, we conjecture that the effectiveness of our \ns\ prompting relies more on the inherent hallucination of the model rather than its size. In other words, the higher the model is hallucinating, the more effective our approach is likely to be.

\subsection{Null-Shot CoT Prompting}\label{sec:ablation:cot}
As 0CoT prompting shows promising increases in performance, we conduct an experiment to investigate if modifying the original 0CoT phrase to include an instruction similar to \ns\ prompting will increase performance or not. The modified 0CoT phrase to incorporate \ns\ prompting, named \nscot\ prompting, is shown in \Cref{sec:cot_phrase}. Similar to 0CoT prompting, we decide to place the \nscot\ phrase at the end of the task instruction and input. We conduct an experiment following the description in \Cref{sec:experiments}, but use 0CoT prompting as a baseline instead of zero-shot prompting. The results from the experiment are shown in \autoref{tab:cot_results} in \Cref{sec:cot_phrase}.

We observe that \nscot\ prompting rarely maintains effectiveness over 0CoT prompting. This could be due to the fact that both prompting approaches require the models to reason and explain in steps, and our \nscot\ prompting may hinder the abilities of the models to reason, resulting in subpar performance compared to 0CoT prompting. Alternatively, we can formulate one observation: eliciting the models to reason also decreases the chances of hallucination.

However, we also perceive that \nscot\ prompting is effective for PaLM 2 in reading comprehension as well as adversarial commonsense reasoning tasks and the Gemini Pro models in the AQuA-RAT dataset. As discussed earlier, base models tend to hallucinate more. By utilizing \nscot\ with long-context inputs, as in RACE datasets and a task that require long reasoning steps, as in WinoGrande, it may guide the models' hallucination in the correct path to the answer. This could also be the reason for the increased performance in the commonsense reasoning and reading comprehension tasks in PaLM 2 for Chat as well. As for the Gemini Pro models, we believe multiple choices in arithmetic questions may intervene with reasoning abilities.

We also note that \nscot\ prompting is very effective for GPT-4 Turbo, previously discussed in \Cref{sec:results:gpt} as a model that seems to have less hallucination, in WinoGrande dataset. The sudden performance increase may suggest that this task likely require both reasoning (from the 0CoT part) and null examples (from the \ns\ part), i.e., easier to elicit hallucination. In contrast, Gemini Pro performs better at handling hallucination in StrategyQA and TriviaQA. As previously discussed, the models are better at detecting non-factual content in the \nscot\ phrase and decline to continue the task. Overall, strong models seem to suffer more from hallucination when using \nscot\ prompting for the AquA-RAT dataset.

\subsection{Positions of Null-Shot Phrase}\label{sec:ablation:positions}
To determine the effects of the position of the \ns\ phrase, we conduct an experiment following what is described in \Cref{sec:experiments}. However, we compare between placing the phrase before the task instruction, as in our original experiment, and at the end of the prompt, similar to what was done in \Cref{sec:ablation:cot}. We only use the GPT-3.5 Turbo model to reduce the cost of the experiment. We compare the obtained performance against the same zero-shot prompting baseline as described in our main experiment. The results are shown in \autoref{tab:position_results} in \Cref{sec:null_shot_ablation_studies}.

We observe that placing the \ns\ phrase at the beginning shows superior effectiveness across datasets, except for GSM8K, where models are required to produce arbitrary numeric answers. We argue that this is due to the fact that placing content at the beginning exhibits stronger conditional strength for these models to rely on for output generation. This phenomenon has also been mentioned in another study, where tokens at the beginning of the prompt have been given more importance compared to the end of the prompt \citep{Liu:2023lost}.

\subsection{Prompt Components}\label{sec:ablation:components}
To assess the contribution of each component in the \ns\ phrase, we conduct an experiment similar to the one described in \Cref{sec:experiments}, but only use GPT-3.5 Turbo to save the cost. We decompose our \ns\ phrase into two main components: ``Look at examples in the ``Examples'' section'' and ``utilize examples and information from that section.'' We prepare three additional variants of the \ns\ phrase, as shown in \Cref{sec:null_shot_ablation_studies}. \texttt{v1} and \texttt{v2} remove the first and second components, respectively; and \texttt{v3} removes both components. Results from the experiment are shown in \autoref{tab:components_results} in \Cref{sec:null_shot_ablation_studies}. We notice that removing both components, as in \texttt{v3}, reduces the effectiveness of \ns\ prompting on all datasets compared to the full \ns\ phrase. Thus, simply instructing the model to perform the task by looking into the null section is insufficient.

We also find that, on the majority of tasks except for arithmetic reasoning and closed-book question answering, \texttt{v2} shows the most prominent performance. Therefore, the first component instructing the model to \textit{look} into the imaginary section plays an important role. However, for the arithmetic reasoning task, we find that \texttt{v1} is most effective, so instructing the model to \textit{utilize} examples and information is crucial for arithmetic tasks. For the closed-book question answering task, both components are required, as can be seen that our full \ns\ phrase provides the best performance, i.e., it requires both \textit{look} and \textit{utilize} instructions. Overall, the full \ns\ phrase may provide the best balance as it encompasses all of the components, making it suitable across tasks.

\section{Conclusion}\label{sec:conclusions}
We propose \ns\ prompting, making use of hallucination in LLMs to improve performance across tasks. Our experiments show the effectiveness of our approach, especially in models with higher innate hallucination, specifically the base models. We also discuss the possibility of utilizing \ns\ prompting for hallucination detection, which can repurpose existing natural language benchmarks for this purpose and does not require task-specific hallucination detection datasets, which are more costly to construct.

We also provide a discussion on the LLMs' behaviors elicited through \ns\ prompting. Furthermore, we conduct various ablation studies exploring scaling effects, reasoning variants of \ns\ prompting, effects of positioning the \ns\ phrase, and the performance contribution of each component in the phrase. Future studies should explore the possibilities of utilizing this approach for detecting hallucination in LLMs and integrating this approach with other PE techniques.

\section*{Limitations}\label{sec:limitations}
While the study that introduced 0CoT prompting \citep{Kojima:2022} used a two-stage prompting approach for improved result extraction, we do not utilize this approach in our study to reduce costs, which may result in some cases of unsuccessful result extraction. However, we compensate it with very flexible output extraction scripts instead (cf. \Cref{sec:datasets}). As stated in \Cref{sec:llms}, we only utilize the 7B version of Llama 2 models, and implications made in this study may change with larger variants of Llama 2. However, we believe that it likely does not change our conclusions about the relationship between innate hallucination and increased performance across tasks.

\section*{Ethics Statement}\label{sec:ethics}
Similar to general use cases of LLMs, our approach is likely to suffer from dataset poisoning \citep{Wallace:2021} as polluted datasets may increase the performance of our approach at the cost of increased hallucination in LLMs. Furthermore, we are unsure about the null examples that models envision during their output generation. Thus, it may retrieve biased, harmful, or toxic content and may lead to the reproduction of such content in the generated outputs. We also note that it is possible to use \ns\ prompting or a modified version of the prompting to avoid harmless and helpful aligned behaviors or other safety mechanisms built into the models and cause jailbreaking \citep{Wei:2023jailbroken}. Finally, as we have a limited understanding of the inner workings of LLMs in general, which is an active area of research, utilizing \ns\ prompting may lead to unexpected behaviors.

\bibliography{null_shot}

\begin{appendices}
\appendix
\section{Related Work}\label{sec:related_work}
\noindent\textbf{Hallucination in LLMs:} Various studies have explored hallucination in LLMs, i.e., behaviors when models provide conflicting information in their outputs \citep{Zhao:2023}. Attempts have been made to reduce hallucination in LLMs across steps in model development, e.g., pre-training \citep{Touvron:2023}, fine-tuning \citep{Bai:2022, Touvron:2023}, and inferencing \citep{Dhuliawala:2023, Li:2023}. These efforts are propelled by the development of various benchmarks for hallucination \citep{Lin:2022, Li:2023}. While it is crucial to reduce hallucination in LLMs, our study proposes that we can exploit these hallucination in LLMs to achieve greater performance across tasks and also utilize this approach for evaluating hallucination in LLMs.

\noindent\textbf{PE:} PE is a  field focused on improving the performance of LLMs through structuring inputs provided to these models, i.e., prompts. Many prompting approaches have been proposed over the years, e.g., few-shot prompting \citep{Brown:2020}, CoT prompting \citep{Wei:2022}, and 0CoT prompting \citep{Kojima:2022}. Many variants of CoT prompting have also been proposed, with their focus either on the \textit{chain}, e.g., chain-of-note \citep{Yu:2023}, CoVe \citep{Dhuliawala:2023}, and chain-of-code \citep{Li:2023chain} prompting. Another line of research focuses on the \textit{thought}, such as tree-of-thought \citep{Yao:2023tree}, graph-of-thought \citep{Besta:2023}, and everything-of-thought \citep{Ding:2023} prompting. While we share similarities with few-shot prompting in utilizing examples and other chain and thought facilities of PE in eliciting longer responses from LLMs, our approach utilizes hallucination in LLMs to use examples that exist within the model. Furthermore, to the best of our knowledge, we are the first to propose PE for hallucination exploitation.

\section{Datasets}\label{sec:datasets}
\autoref{fig:question_formatting} displays the format of task instructions and inputs for the datasets. This format is inspired by the procedure done in the 0CoT prompting study \citep{Kojima:2022}. Choices and additional context are only provided in the prompts when applicable. All included datasets are in English. Additional details on the chosen testing set and the number of records are presented in \autoref{tab:datasets}.

\begin{figure}[tbp]
\begin{mainbox}{}
\textcolor{purple}{Additional context, e.g., article, context, and hypothesis}\\
Question: \textcolor{purple}{\{question\}}\\
Choices: \textcolor{purple}{\{choices\}}\\
Answer:
\end{mainbox}
\caption{The task instruction and task input format used for the experiments.}
\label{fig:question_formatting}
\end{figure}

\begin{table}[tbp]
\centering\footnotesize
\begin{tabular}{rcccc}
\hline
\textbf{Dataset} & \textbf{Task} & \textbf{Test split} & \textbf{Count} & \textbf{Ans.}\\
\hline
AQuA-RAT & AR & \texttt{test} & 254 & MC\\
GSM8K & AR & \texttt{test} & 1319 & Num.\\
StrategyQA & CR & \texttt{test} & 2290 & BC\\
WinoGrande & CR & \texttt{dev} & 1267 & BC\\
RACE-m & RC & \texttt{middle-test} & 1436 & MC\\
RACE-h & RC & \texttt{high-test} & 3498 & MC\\
ANLI & NLI & \texttt{R3-test} & 1200 & MC\\
TriviaQA & CQA & \texttt{Wikipedia} & 1000$^*$ & Text\\
\hline
\end{tabular}
\caption{Details of each dataset. \textbf{Test split} shows the split used for evaluations in this study, while \textbf{Count} shows the number of included samples. For the \textbf{Task}, \textit{AR}: Arithmetic Reasoning, \textit{CR}: Commonsense Reasoning, \textit{RC}: Reading Comprehension, \textit{NLI}: Natural Language Inference, \textit{CQA}: Closed-book Question Answering. The \textbf{Ans.} denotes the type of the expected answer, where \textit{BC} represents binary choices, \textit{MC} represents multiple choices, \textit{Num.} represents an arbitrary number answer, and \textit{Text} represents a free-text answer.\\\footnotesize{$^*$We downsampled TriviaQA to only 1000 records to save budget.}}
\label{tab:datasets}
\end{table}

We note that AQuA-RAT, WinoGrande, and TriviaQA are under the Apache License, Version 2.0. GSM8K and StrategyQA are under the MIT License. RACE datasets are available for non-commercial research purposes only. ANLI is under the Creative Commons Attribution-NonCommercial 4.0 International License. TriviaQA used in our study is downsampled using the standard random sampling function in Python with a fixed seed of 42. We also note that the datasets may include names of individuals collected from the internet, i.e., publicly available facts about a person but not in an offensive way. The following list shows the sources of data we used for this study.

\begin{itemize}
    \item AQuA-RAT: \url{https://github.com/google-deepmind/AQuA}
    \item GSM8K: \url{https://github.com/openai/grade-school-math}
    \item StrategyQA: \url{https://github.com/google/BIG-bench/tree/main/bigbench/benchmark_tasks/strategyqa}
    \item WinoGrande: \url{https://winogrande.allenai.org}
    \item RACE: \url{https://www.cs.cmu.edu/~glai1/data/race/}
    \item ANLI: \url{https://github.com/facebookresearch/anli}
    \item TriviaQA: \url{https://nlp.cs.washington.edu/triviaqa/}
\end{itemize}

We also develop output extraction scripts for all datasets. For datasets with choices, we look for patterns of choices in the responses. First, if the response generated from a model is an uppercase character, we treat that as the final answer. For example, if a model responded with ``A'' and if we have ``A'' as one of our choices, ``A'' will be treated as the final answer. In other cases, we first attempt to match a pattern of an uppercase character choice followed by a parenthesis, e.g., ``A)''. Then we try to match a pattern of ``answer is'', where we treat the first uppercase character choice after the pattern as the final answer. For example, if a response contains ``So, the answer is A)'', ``A'' will be extracted as the final answer.

For all patterns, we attempt to match on the last line of the model's output first. If unsuccessful, we then try to match the first line of the model output. These heuristics are based on our observation that models are likely to provide the conclusive answer in the last or first lines, as empirically observed in our pilot study. Failures to match are treated as no answer, as well as in cases where the model returns an empty response.

For datasets without choices, two scenarios are considered. The first scenario is when the answer is a number. In this case, we treat the first number found on the last or first line as the final answer. This is in a similar spirit to a previous study \citep{Kojima:2022}. The second scenario is when the answer is free text. In this case, we first lowercase the response and the label. Then we check if the label exists in the response or not.

\section{LLMs}\label{sec:llms}
All LLMs in this study are utilized in a deterministic setup, i.e., we set the sampling temperature to 0 and provide a fixed random seed when applicable. Therefore, we only interact with the model once for each record of the dataset given a prompting approach. Any additional settings, including safety, are left to default. For chat models/pipelines, we always start with an empty context history, with the prompt as the first user message. Six LLMs included in the main experiment are PaLM 2 (\texttt{text-bison-001}), PaLM 2 for Chat (\texttt{chat-bison-001}), Gemini Pro (\texttt{gemini-pro}) via \texttt{generateContent} method, Gemini Pro (Chat) (\texttt{gemini-pro} via \texttt{start\_chat} method), GPT-3.5 Turbo (\texttt{gpt-3.5-turbo-1106}), and GPT-4 Turbo (\texttt{gpt-4-1106-preview}). We choose these models for our experiments as they offer APIs to access the models without the need to prepare our own infrastructure for running the models. Furthermore, all of these models are relatively large and are utilized in many real-world products and scenarios.

PaLM 2 and PaLM 2 for Chat serve as a comparison for models from the same family, where one model is possibly a base model and the other one is potentially a chat fine-tuned variant for chat conversation. This could further give us a way to assess the effectiveness of the proposed prompting between these two types of LLMs and the importance of chat fine-tuning. Similarly, GPT-3.5 Turbo and GPT-4 Turbo are also chosen to assess these instruction-aligned models within the same family, where the subsequent version of the same model family is possibly relatively larger in both parameter size and training data. This could provide insights into the effects of scaling models further. We include Gemini Pro because its performance is likely positioned between that of GPT-3.5 Turbo and GPT-4 Turbo. All of these six aforementioned LLMs are utilized via their respective API-wrapper Python libraries\footnote{GPT-3.5 Turbo and GPT-4 Turbo: \url{https://github.com/openai/openai-python}\\\indent PaLM 2, PaLM 2 for Chat, Gemini Pro, and Gemini Pro (Chat): \url{https://github.com/google/generative-ai-python}}.

Finally, Llama 2 7B\footnote{\url{https://huggingface.co/meta-llama/Llama-2-7b-hf}} and Llama 2 7B Chat\footnote{\url{https://huggingface.co/meta-llama/Llama-2-7b-chat-hf}} are included for the scaling analysis where we attempt to evaluate \ns\ prompting on smaller models. Due to resource constraints, we are only able to evaluate using the smallest variants of Llama 2 models (7B), despite the existence of the 13B and 70B variants. Both Llama 2 models are utilized via Hugging Face's \texttt{transformers}\footnote{\url{https://github.com/huggingface/transformers/}} pipelines, i.e., the \texttt{text-generation} pipeline for Llama 2 and the \texttt{conversational} pipeline for Llama 2 Chat. We note that all models used in our study through APIs are subject to the terms and conditions of API providers, which allow non-commercial research purposes in our study. Llama 2 models are subject to Meta's community license agreement, which permits our use cases.

\section{Null-Shot CoT Phrase}\label{sec:cot_phrase}
This section contains phrases used for the \nscot\ prompting experiment and relative results of the experiment described in \Cref{sec:ablation:cot}. The \nscot\ and 0CoT phrases are shown in Figures \ref{fig:null_shot_cot_phrase} and \ref{fig:0cot_phrase}, respectively. The results from the experiment are presented in \autoref{tab:cot_results}.

\begin{figure}[tbp]
\begin{mainbox}{\nscot\ Phrase}
Look at examples in the ``Examples'' section and utilize examples and information from that section to perform the following task \circlebox{step-by-step}.
\end{mainbox}
\caption{The \nscot\ phrase shares the majority with the \ns\ phrase. However, it has ``step-by-step'', highlighted in yellow, added at the end, inspired by 0CoT prompting.}
\label{fig:null_shot_cot_phrase}
\end{figure}

\begin{figure}[tbp]
\begin{mainbox}{0CoT Phrase}
Let’s think step by step.
\end{mainbox}
\caption{The 0CoT phrase used as a baseline for the experiment detailed in \Cref{sec:ablation:cot}. This phrase is taken from \citet{Kojima:2022}.}
\label{fig:0cot_phrase}
\end{figure}

\begin{table*}[tbp]
\centering\small
\begin{tabular}{rcccccccc}
\hline
\textbf{Model} & \textbf{AQuA} & \textbf{GSM8K} & \textbf{StrategyQA} & \textbf{WinoGrande} & \textbf{RACE-m} & \textbf{RACE-h} & \textbf{ANLI} & \textbf{TriviaQA}\\
\hline
PaLM 2 & -54.44\% & -27.71\% & -5.36\% & \textcolor{mygreen}{18.89\%} & \textcolor{mygreen}{\textbf{16.18\%}} & \textcolor{mygreen}{\textbf{20.16\%}} & \textbf{-0.85\%} & -4.50\%\\
PaLM 2 for Chat & -5.88\% & -7.55\% & \textcolor{mygreen}{\textbf{14.75\%}} & -4.49\% & \textcolor{mygreen}{0.49\%} & \textcolor{mygreen}{0.79\%} & -2.49\% & -2.12\%\\
GPT-3.5 Turbo & -3.42\% & -4.00\% & -13.21\% & -10.96\% & -3.34\% & \textcolor{mygreen}{0.15\%} & -46.94\% & \textcolor{mygreen}{\textbf{0.75\%}}\\
GPT-4 Turbo & \textcolor{mygreen}{2.08\%} & \textbf{-1.12\%} & -2.34\% & \textcolor{mygreen}{\textbf{24.77\%}} & -5.64\% & -8.56\% & -8.79\% & -1.04\%\\
Gemini Pro & \textcolor{mygreen}{\textbf{8.47\%}} & -9.99\% & -98.42\% & -99.62\% & -3.02\% & -1.06\% & -7.07\% & -98.54\%\\
Gemini Pro (Chat) & \textcolor{mygreen}{8.06\%} & -11.66\% & -98.42\% & -99.62\% & -2.32\% & -0.73\% & -8.01\% & -98.55\%\\
\hline
\end{tabular}
\caption{This table presents relative results indicating the performance changes when using \nscot\ prompting compared to the 0CoT prompting baseline.}
\label{tab:cot_results}
\end{table*}

\section{Null-Shot Phrase's Positioning and Null-Shot Phrase Variants}\label{sec:null_shot_ablation_studies}
In this section, we present a result table, \Cref{tab:position_results}, of the experiment described in \Cref{sec:ablation:positions} to determine the position placement of the \ns\ phrase. We also present prompt variants used for the experiment described in \Cref{sec:ablation:components} to determine the effects of each component in Figures \ref{fig:null_shot_v1}, \ref{fig:null_shot_v2}, and \ref{fig:null_shot_v3}, and their results in \Cref{tab:components_results}.

For the prompt variants, the original \ns\ phrase is shown in \autoref{fig:null_shot_phrase} and has been broken down into two main components, as shown in \autoref{fig:null_shot_components}. The first variant, \texttt{v1}, removes the first component. The second variant, \texttt{v2}, removes the second component. Finally, the third variant, \texttt{v3}, removes both components. The full phrases are shown in Figures \ref{fig:null_shot_v1}, \ref{fig:null_shot_v2}, and \ref{fig:null_shot_v3}.

\begin{figure}[tbp]
\begin{mainbox}{\ns\ Phrase: Components}
\circlebox{Look at examples in the ``Examples'' section} and \circlebox[mydarkgreen]{utilize examples and information from} \circlebox[mydarkgreen]{that section} to perform the following task.
\end{mainbox}
\caption{The \ns\ phrase with the first components highlighted in yellow and the second components highlighted in green.}
\label{fig:null_shot_components}
\end{figure}

\begin{figure}[tbp]
\begin{mainbox}{\ns\ Phrase: First Variant  (v1)}
Utilize examples and information from the ``Examples'' section to perform the following task.
\end{mainbox}
\caption{The first variant of \ns\ phrase with the first component removed.}
\label{fig:null_shot_v1}
\end{figure}

\begin{figure}[tbp]
\begin{mainbox}{\ns\ Phrase: Second Variant  (v2)}
Look at examples in the ``Examples'' section and perform the following task.
\end{mainbox}
\caption{The second variant of \ns\ phrase with the second component removed.}
\label{fig:null_shot_v2}
\end{figure}

\begin{figure}[tbp]
\begin{mainbox}{\ns\ Phrase: Third Variant  (v3)}
Perform the following task as demonstrated in the ``Examples'' section.
\end{mainbox}
\caption{The third variant of \ns\ phrase with both components removed.}
\label{fig:null_shot_v3}
\end{figure}

\begin{table*}[tbp]
\centering\small
\begin{tabular}{rcccccccc}
\hline
\textbf{Prompting} & \textbf{AQuA} & \textbf{GSM8K} & \textbf{StrategyQA} & \textbf{WinoGrande} & \textbf{RACE-m} & \textbf{RACE-h} & \textbf{ANLI} & \textbf{TriviaQA}\\
\hline
\ns & \textcolor{mygreen}{\textbf{33.94\%}} & \textcolor{mygreen}{15.19\%} & \textcolor{mygreen}{\textbf{3.14\%}} & \textbf{-1.84\%} & \textbf{-1.79\%} & \textbf{-1.19\%} & \textbf{-3.61\%} & \textcolor{mygreen}{\textbf{1.23\%}}\\
\ns-after & \textcolor{mygreen}{30.28\%} & \textcolor{mygreen}{\textbf{19.20\%}} & -6.21\% & -69.08\% & -4.81\% & -4.62\% & -46.47\% & -3.09\%\\
\hline
\end{tabular}
\caption{This table presents relative results indicating the performance changes compared to the zero-shot prompting baseline. We find that placing the phrase before the task instruction and input allows GPT-3.5 Turbo to perform better in most cases.}
\label{tab:position_results}
\end{table*}

\begin{table*}[tbp]
\centering\small
\begin{tabular}{rcccccccc}
\hline
\textbf{Prompting} & \textbf{AQuA} & \textbf{GSM8K} & \textbf{StrategyQA} & \textbf{WinoGrande} & \textbf{RACE-m} & \textbf{RACE-h} & \textbf{ANLI} & \textbf{TriviaQA}\\
\hline
\ns & \textcolor{mygreen}{33.94\%} & \textcolor{mygreen}{15.19\%} & \textcolor{mygreen}{3.14\%} & -1.84\% & -1.79\% & -1.19\% & -3.61\% & \textcolor{mygreen}{\textbf{1.23\%}}\\
\hline
\ns-v1 & \textcolor{mygreen}{\textbf{36.70\%}} & \textcolor{mygreen}{\textbf{16.85\%}} & \textcolor{mygreen}{2.73\%} & -3.95\% & -2.12\% & \-1.75\% & -6.37\% & \textcolor{mygreen}{0.62\%}\\
\ns-v2 & \textcolor{mygreen}{10.09\%} & \textcolor{mygreen}{8.98\%} & \textcolor{mygreen}{\textbf{4.57\%}}& \textcolor{mygreen}{\textbf{1.84\%}} & \textbf{-1.14\%} & \textbf{-0.80\%} & \textcolor{mygreen}{\textbf{0.52\%}} & -0.37\%\\
\ns-v3 & \textcolor{mygreen}{27.52\%} & \textcolor{mygreen}{15.88\%} & \textcolor{mygreen}{1.23\%} & -8.82\% & -1.47\% & -1.43\% & -2.07\% & -0.25\%\\
\hline
\end{tabular}
\caption{This table presents relative results indicating the performance changes compared to the zero-shot prompting baseline. We note that each prompt component has its own strengths. In particular, \texttt{v2}, where the second prompt component is removed, shows the best performance in most datasets.}
\label{tab:components_results}
\end{table*}

\section{Hallucination in LLMs in Relation to Phenomena in Humans}\label{sec:with_human}
In this section, we engage in a broader discussion about the similarities of hallucination in LLMs to three phenomena in humans: déjà vu, fabrication, and confabulation. We select these phenomena as they share some characteristics with LLMs, e.g., false memory, lying, or double firing in the same region of neurons. We discuss déjà vu, fabrication, and confabulation in Sections \ref{sec:with_human:dejavu}, \ref{sec:with_human:fabrication}, and \ref{sec:with_human:confabulation}, respectively.

\subsection{Déjà Vu}\label{sec:with_human:dejavu}
Déjà vu \citep{Brown:2023} in humans is a phenomenon where one believes that they have a memory of a certain situation before. For example, they may feel like they have visited a place before, but in fact, this is their first time there. The reasons behind this phenomenon are still inconclusive for humans \citep{Brown:2023}, with many streams of research pursuing explanations.

We believe that one potential reason behind the increased performance of LLMs could be due to a similar phenomenon. LLMs may believe that they have seen a situation before, while in fact, they have never seen such a situation during their training. In contrast to emergent abilities \citep{Wei:2022emergent} when LLMs are further scaled and they generalize to never-before-seen tasks, this phenomenon is possibly due to LLMs believing that they have seen a particular piece of information before, similar to déjà vu in humans, enabling them to retrieve a memory that may have never existed. If this holds true, it could also be a key factor in utilizing LLMs for better understanding déjà vu in humans.

The most similar type of explanation for this behavior in LLMs to déjà vu in humans, in our opinion, is the attentional framework \citep{Brown:2023}, where our brain may process information in two passes, with the first pass being inattention and the second one being full attention. The matching between an experience of the second pass to the first pass makes us subconsciously feel like we had this experience before. In LLMs, it may be possible that \ns\ prompting elicits similar behaviors of firing through same regions of attention weights, in a loose sense, twice, making the model hallucinate the null section. The interpretation of this could be due to a resurfacing of similar probability distributions of tokens during LLMs' decoding process.

\subsection{Fabrication}\label{sec:with_human:fabrication}
Another perspective to consider is fabrication. As humans, we fabricate, i.e., lie about facts, stories, experiences, and more \citep{Saxe:1911}. We fabricate for various purposes, such as protecting our loved ones from harsh truths, maintaining harmony among peers, or taking advantage of a situation through fabricated stories. Considering that LLMs have been trained on large corpora containing a massive amount of human-generated content \citep{Zhao:2023}, these models may learn these kinds of behaviors through their training data. Alternatively, it could be due to the fact that the training corpora may contain conflicting data, leading to hallucinatory behaviors of LLMs. Fabricating the null ``Examples'' section as instructed in \ns\ prompting, is potentially done because the model wants to maintain comfort or gain favors, i.e., ``sycophancy'', with users \citep{perez:2023}.

While fabrication in this sense may sound acceptable, these behaviors of fabricating facts can be exploited in malicious attempts by making the models fabricate false information, strengthening confirmation bias \citep{Nickerson:1998} instead of providing truthful and objective information. This kind of hallucination can be harmful, and while we propose \ns\ prompting to increase performances of LLMs by exploiting inherent hallucination, we posit that a better understanding and mitigation of hallucination in LLMs should render our approach less effective. This means that LLMs are less prone to hallucination and can provide more truthful information. That is why we also posit that \ns\ prompting shows the possibility of uses in hallucination detection as well.

\subsection{Confabulation}\label{sec:with_human:confabulation}
Related to déjà vu and fabrication is confabulation. Confabulation in humans is an ``honest lying'' \cite{Berrios:1998} where a person retains a false memory and believes that such a memory is true \cite{Fotopoulou:2008}. Similarly, as we observe from the results, LLMs may honestly believe that such a section exists when prompted with \ns\ prompting and try to produce results in accordance with the instruction in the prompt. In humans, provoked confabulation \cite{Schnider:1996,Francis:2022} directly \textit{prompts} a person with a question or conversation related to a false memory. This type of confabulation can also be regarded as the same as what \ns\ phrase \textit{prompts} LLMs.

While confabulation is regarded as a neuropsychiatric disorder usually following brain damage, comprehensive causes of this disorder remain inconclusive \cite{Berrios:1998,Francis:2022}. Further investigation and understanding in LLMs for the origin of their hallucination may also shed some light and aid in discovering causes of confabulation in humans. Nevertheless, confabulation, both in humans and LLMs, is generally regarded as an undesired behavior, and various studies have been explore intervention/mitigation approaches \cite{Francis:2022,Zhang:2023}. Finally, we acknowledge that some studies use confabulation in place of hallucination for LLMs \cite{Shanahan:2023,Rawte:2023}. Whether which term is more suitable to describe this category of behaviors in LLMs remains inconclusive for the field and is an open question.

\section{Empirical Results of Expected Behaviors When Using Null-Shot Prompting}\label{sec:empirical_results_less}
This section contains qualitative examples generated by either GPT-4 Turbo or Gemini Pro from our main experiment and from ChatGPT web version. When we utilize \ns\ prompting, these models may inform users about the unavailability of the ``Examples'' section. This demonstrates a less hallucinatory behavior and may be preferred in scenarios where, for example, users unintentionally forget to provide the stated section in the prompt but intend to include it. Through these examples, we find that only GPT-4 Turbo and Gemini Pro have the ability to inform users about its inaccessibility to the instructed null ``Examples'' section. This behavior exhibits less hallucination compared to other models, specifically, context-conflicting hallucination. The numbers of instances for each dataset where this event occurred, i.e., GPT-4 Turbo starts the response with ``I'm sorry, but'' or Gemini Pro generates something similar to ``The provided context does not contain'', are presented in Tables \ref{tab:no_access_gpt_4}, \ref{tab:no_access_gemini_pro}, and \ref{tab:no_access_gemini_pro_chat}, for GPT-4 Turbo, Gemini Pro, and Gemini Pro (Chat), respectively. Additional qualitative examples are shown in Figures \ref{fig:empirical_example_gpt_4_gsm8k}, \ref{fig:empirical_example_gpt_4_strategyqa}, \ref{fig:empirical_example_gpt_4_triviaqa}, \ref{fig:empirical_example_gemini_pro_strategyqa}, and \ref{fig:empirical_example_gemini_pro_chat_triviaqa}.

\begin{table}[tbp]
\centering\small
\begin{tabular}{cccc}
\hline
\textbf{AQuA} & \textbf{GSM8K} & \textbf{StrategyQA} & \textbf{WinoGrande}\\
\hline
0.39\% (1) & 0.08\% (1) & 53.89\% (1234) & 20.52\% (260)\\
\hline
\end{tabular}
\begin{tabular}{cccc}
\hline
\textbf{RACE-m} & \textbf{RACE-h} & \textbf{ANLI} & \textbf{TriviaQA}\\
\hline
0\% (0) & 0.06\% (2) & 0\% (0) & 7.9\% (79)\\
\hline
\end{tabular}
\caption{Number of instances when GPT-4 Turbo's response includes a phrase informing the user about the unavailability of the instructed ``Examples''.}
\label{tab:no_access_gpt_4}
\end{table}

\begin{table}[tbp]
\centering\small
\begin{tabular}{cccc}
\hline
\textbf{AQuA} & \textbf{GSM8K} & \textbf{StrategyQA} & \textbf{WinoGrande}\\
\hline
0.39\% (1) & 1.29\% (17) & 26.33\% (603) & 5.45\% (69)\\
\hline
\end{tabular}
\begin{tabular}{cccc}
\hline
\textbf{RACE-m} & \textbf{RACE-h} & \textbf{ANLI} & \textbf{TriviaQA}\\
\hline
0\% (0) & 0.2\% (7) & 0\% (0) & 64.7\% (647)\\
\hline
\end{tabular}
\caption{Number of instances when Gemini Pro's response includes a phrase informing the user about the unavailability of the instructed ``Examples''.}
\label{tab:no_access_gemini_pro}
\end{table}

\begin{table}[tbp]
\centering\small
\begin{tabular}{cccc}
\hline
\textbf{AQuA} & \textbf{GSM8K} & \textbf{StrategyQA} & \textbf{WinoGrande}\\
\hline
0.39\% (1) & 1.21\% (16) & 26.24\% (601) & 5.45\% (69)\\
\hline
\end{tabular}
\begin{tabular}{cccc}
\hline
\textbf{RACE-m} & \textbf{RACE-h} & \textbf{ANLI} & \textbf{TriviaQA}\\
\hline
0\% (0) & 0.23\% (8) & 0\% (0) & 64.8\% (648)\\
\hline
\end{tabular}
\caption{Number of instances when Gemini Pro (Chat)'s response includes a phrase informing the user about the unavailability of the instructed ``Examples''.}
\label{tab:no_access_gemini_pro_chat}
\end{table}

\subsection{Context-Conflicting Hallucination Detection Ability of GPT-4 Turbo}\label{sec:empirical_results_less:gpt_4}
As can be observed from \autoref{tab:no_access_gpt_4}, GPT-4 Turbo is less prone to hallucination when using our \ns\ prompting in StrategyQA and WinoGrande compared to other datasets, despite the fact that our \ns\ prompting elicits context-conflicting hallucination. Typically, commonsense reasoning requires a use of implicit reasoning steps \citep{Geva:2021} or world knowledge \citep{Levesque:2012}; performing this task may induced the model to utilize its associated weights of various reasoning types required by each question in the task. The use of reasoning may resulted in reduced hallucination; in our case, the model is better at detecting conflicting instructions. This observation is aligned with a previous study \citep{Dhuliawala:2023} which showed that reasoning could reduce LLMs' hallucination.

TriviaQA is another task where the model shows its ability to detect hallucination compared to the rest of the dataset. This could be due to the fact that trivia questions may require additional knowledge, eliciting GPT-4 Turbo to search the Internet or retrieve information from external sources, as this approach is common for this task \citep{Yasunaga:2021,Schick:2023}. As GPT-4 might attempt to access these additional sources but could not, the model responded with the unavailability of the section.

On the other hand, GPT-4 Turbo did not inform users in arithmetic reasoning, reading comprehension, and NLI tasks. These tasks may have different characteristics that may not encourage the model to reason through words. For example, the reading comprehension task may require a general level of reasoning. However, with its long-context nature, this may prohibit GPT-4 Turbo from reasoning and easily distract the model via our \ns\ phrase, as we instructed the model to further look into something that sounds promising to exist given the long context. For arithmetic reasoning, numbers, calculations, and mathematical symbols may distract the model from paying attention to detect the conflict in the prompt, i.e., activated different areas of attentions. As for NLI, assessing a given hypothesis against a provided context may not be enough to elicit the reasoning level necessary to detect conflicts in prompts.

\subsection{Context-Conflicting Hallucination Detection Ability of Gemini Pro}\label{sec:empirical_results_less:gemini_pro}
Similar to what can be observed with GPT-4 Turbo, Gemini Pro is able to detect context-conflicting hallucination in datasets, as shown in Tables \ref{tab:no_access_gemini_pro} and \ref{tab:no_access_gemini_pro_chat}. In contrast to GPT-4 Turbo, we observe a noticeable rate of over half of the generated responses for TriviaQA, but not WinoGrande, containing an informing statement that the instruction to utilize information or examples from the null ``Examples'' section is incorrect. We note that both Gemini Pro and Gemini Pro (Chat) share a highly similar pattern across datasets, likely due to them being a similar model.

We observe that arithmetic reasoning and reading comprehension tasks, coupled with \ns\ prompting, lower the ability of the models to reason and detect hallucination, the same as with GPT-4 Turbo. Therefore, we conjecture that this is due to the nature of the tasks, which involve heavy numerical values and long contexts in general. We prompt future studies to design hallucination detection methods incorporating this insight during the development of hallucination detection datasets. Interestingly, TriviaQA is where the models shine the most, which is consistent with a report on Gemini where the authors implemented instruction-tuning approaches aiming at reducing incorrect information generation in closed-book question answering tasks \citep{Gemini:2023}.

\subsection{Inability of Other LLMs to Detect Hallucination}\label{sec:empirical_results_less:others}
One potential reason why other models could not detect context-conflicting hallucination when using our \ns\ prompting could be due to the fact that these models are smaller compared to GPT-4 Turbo or Gemini Pro. In a previous study, it showed that smaller models may exhibit fewer reasoning capabilities and more hallucinated behaviors \citep{Wei:2022}. Therefore, even Llama 2 models were trained using a process designed to minimize hallucination, they exhibited more hallucinatory behaviors, likely due to the lack of scale.

As for PaLM 2 models and GPT-3.5 Turbo, it is unclear how their scale is comparable to GPT-4 Turbo or Gemini Pro due to a lack of public report. Nevertheless, it is worth noting that GPT-3.5 utilized through the ChatGPT website exhibits better responses in informing users about the inaccessibility of the null section. An example of an interaction with GPT-3.5 through the ChatGPT website is shown in \autoref{fig:empirical_example_chatgpt}. The inconsistency in behaviors between GPT-3.5 utilized via the website and GPT-3.5 Turbo utilized via the API could possibly be due to the constant updates behind the scenes of the web version, which is potentially powered by a newer model.

\begin{figure*}[tbp]
\begin{tcolorbox}[colback=myblue!5!white,colframe=myblue]
Look at examples in the ``Examples'' section and utilize examples and information from that section to perform the following task.\\
Question: How much does it cost you for lunch today at Subway if you pay \$40 for a foot-long fish sub and thrice as much for a six-inch cold-cut combo sub?\\
Answer:
\tcblower
\textbf{I'm sorry, but there seems to be a misunderstanding. I don't have access to an ``Examples'' section or any other external content.} However, I can help you calculate the cost based on the information you've provided.\\
\\
If you paid \$40 for a foot-long fish sub at Subway, and a six-inch cold-cut combo sub costs thrice as much, then the cost of the six-inch cold-cut combo sub would be:\\
\\
3 times \$40 = \$120\\
\\
So, the total cost for lunch at Subway would be the cost of the foot-long fish sub plus the cost of the six-inch cold-cut combo sub:\\
\\
\$40 (foot-long fish sub) + \$120 (six-inch cold-cut combo sub) = \$160\\
\\
Therefore, it would cost you \textcolor{mygreen}{\textbf{\$160}} for lunch today at Subway based on the prices you've provided. However, please note that these prices seem unusually high for Subway sandwiches, and it's likely that there's been a mistake or miscommunication regarding the actual costs.
\end{tcolorbox}
\caption{An output generated using \ns\ prompting with GPT-4 Turbo for GSM8K dataset.}
\label{fig:empirical_example_gpt_4_gsm8k}
\end{figure*}

\begin{figure*}[tbp]
\begin{tcolorbox}[colback=myblue!5!white,colframe=myblue]
Look at examples in the ``Examples'' section and utilize examples and information from that section to perform the following task.\\
Question: Is one blast from double-barreled shotgun likely to kill all squid brains?\\
Choices: A) True, B) False\\
Answer:
\tcblower
\textbf{I'm sorry, but as an AI, I don't have access to external content, including an ``Examples'' section you're referring to. Therefore, I cannot directly utilize examples and information from that section to answer your question.}\\
\\
However, I can provide a general answer based on common knowledge. A double-barreled shotgun is a powerful firearm that can cause significant damage. Whether one blast could kill all squid brains would depend on several factors, including the size and number of the squid, the distance, the spread of the shot, and the specific ammunition used. In a realistic scenario, it is unlikely that one blast would kill multiple squid if they are not in very close proximity to each other, as the shot spread would disperse over a distance, reducing its lethality per squid. Therefore, without specific context, the most accurate answer would be:\\
\\
Answer: \textcolor{mygreen}{\textbf{B) False}}
\end{tcolorbox}
\caption{An output generated using \ns\ prompting with GPT-4 Turbo for StrategyQA dataset.}
\label{fig:empirical_example_gpt_4_strategyqa}
\end{figure*}

\begin{figure*}[tbp]
\begin{tcolorbox}[colback=myblue!5!white,colframe=myblue]
Look at examples in the ``Examples'' section and utilize examples and information from that section to perform the following task.\\
Question: `When Will You Marry?' became the world's most expensive work of art when it was sold\\
Answer:
\tcblower
\textbf{I'm sorry, but as an AI, I don't have the ability to look at sections or examples outside of the text provided to me in our conversation.} However, I can tell you that the painting ``When Will You Marry?'' by \textcolor{mygreen}{\textbf{Paul Gauguin}} was sold for a record price of nearly \$300 million in 2015. At that time, it was reported to be one of the most expensive works of art ever sold. Please note that the art market is dynamic, and records for the most expensive works of art can change over time.
\end{tcolorbox}
\caption{An output generated using \ns\ prompting with GPT-4 Turbo for TriviaQA dataset.}
\label{fig:empirical_example_gpt_4_triviaqa}
\end{figure*}

\begin{figure*}[tbp]
\begin{tcolorbox}[colback=myblue!5!white,colframe=myblue]
Look at examples in the ``Examples'' section and utilize examples and information from that section to perform the following task.
Question: Did Metallica band members cutting their hair hurt their sales?
Choices: A) True, B) False
Answer:
\tcblower
\textbf{The provided context does not contain any information about Metallica band members cutting their hair or the impact it had on their sales. Therefore, I cannot answer this question.}
\end{tcolorbox}
\caption{An output generated using \ns\ prompting with Gemini Pro for StrategyQA dataset.}
\label{fig:empirical_example_gemini_pro_strategyqa}
\end{figure*}

\begin{figure*}[tbp]
\begin{tcolorbox}[colback=myblue!5!white,colframe=myblue]
Look at examples in the ``Examples'' section and utilize examples and information from that section to perform the following task.
Question: ``Which actor, best known for his role in ````The Godfather'''' series of films, shares his name with a member of TV's ````Dragon's Den''''?''
Answer:
\tcblower
\textbf{The context does not mention anything about actors in ``The Godfather'' series of films or members of TV's ``Dragon's Den'', so I cannot answer this question from the provided context.}
\end{tcolorbox}
\caption{An output generated using \ns\ prompting with Gemini Pro (Chat) for TriviaQA dataset.}
\label{fig:empirical_example_gemini_pro_chat_triviaqa}
\end{figure*}

\begin{figure*}[tbp]
\begin{tcolorbox}[colback=myblue!5!white,colframe=myblue]
Look at examples in the ``Examples'' section and utilize examples and information from that section to perform the following task.\\
Question: The A57 road runs from Liverpool to which English town or city?\\
Answer:
\tcblower
I'm sorry for any confusion, but \textbf{I don't have direct access to specific sections, including an ``Examples'' section}, as I don't have the ability to browse external content or specific webpages. However, I can provide information based on my training data up to January 2022.\\
\\
As of my last update, the A57 road runs from Liverpool to \textcolor{mygreen}{\textbf{Lincoln}}. It's a major road in England that connects these two locations. If there have been any changes or updates after my last training cut-off, I recommend checking a reliable and up-to-date source for the latest information on the A57 road.
\end{tcolorbox}
\caption{An output generated using \ns\ prompting with ChatGPT (web version). An archived version of the conversation can be accessed via \url{https://chat.openai.com/share/96d07393-c685-45a1-a321-72471bc91c41}}
\label{fig:empirical_example_chatgpt}
\end{figure*}

\section{Error Analysis}\label{sec:error_analysis}
We investigate failure cases of the main experiment. In particular, we focus on cases where responses are an empty string due to getting blocked from safety mechanisms built into these models or their APIs; we leave all safety settings to default to imitate real-world scenarios of API usages. We note that these mechanisms, as of writing, only exist within the models used through APIs served by Google which are the PaLM 2\footnote{\url{https://cloud.google.com/vertex-ai/docs/generative-ai/configure-safety-attributes-palm}} models and Gemini Pro\footnote{\url{https://ai.google.dev/docs/safety_setting_gemini}} models. Our further investigations also validate that GPT-3.5 Turbo and GPT-4 Turbo do not have this behavior. \autoref{tab:empty_responses} presents cases where the aforementioned models from Google output empty responses due to being blocked by the security mechanisms.

\begin{table*}[tbp]
\centering
\scriptsize
\begin{tabular}{rcc|cc|cc|cc}
\hline
\multirow{2}{*}{\textbf{Model}} & \multicolumn{4}{c|}{\textbf{AQuA}} & \multicolumn{4}{c}{\textbf{GSM8K}}\\
& \textbf{0S} & \textbf{$\varnothing$S} & \textbf{0CoT} & \textbf{$\varnothing$CoT} & \textbf{0S} & \textbf{$\varnothing$S} & \textbf{0CoT} & \textbf{$\varnothing$CoT}\\
\hline
PaLM 2 & 2.36\% (6) & \textcolor{myred}{0\% (0)} & 2.76\% (7) & \textcolor{myred}{2.36\% (6)} & 4.02\% (53) & \textcolor{myred}{0.38\% (5)} & 1.14\% (15) & \textcolor{myred}{0.91\% (12)}\\
PaLM 2 for Chat & 0.39\% (1) & \textcolor{myred}{0\% (0)} & 0\% (0) & 0\% (0) & 0.23\% (3) & 0.3\% (4) & 0.23\% (3) & 0.23\% (3)\\
Gemini Pro & 1.18\% (3) & \textcolor{myred}{0.39\% (1)} & 0\% (0) & 0.39\% (1) & 3.26\% (43) & \textcolor{myred}{0.53\% (7)} & 1.06\% (14) & \textcolor{myred}{0.45\% (6)}\\
Gemini Pro (Chat) & 0.39\% (1) & 0.39\% (1) & 0\% (0) & 0\% (0) & 3.26\% (43) & \textcolor{myred}{0.3\% (4)} & 0.53\% (7) & \textcolor{myred}{0.38\% (5)}\\
\hline
\end{tabular}

\begin{tabular}{rcc|cc|cc|cc}
\hline
\multirow{2}{*}{\textbf{Model}} & \multicolumn{4}{c|}{\textbf{StrategyQA}} & \multicolumn{4}{c}{\textbf{WinoGrade}}\\
& \textbf{0S} & \textbf{$\varnothing$S} & \textbf{0CoT} & \textbf{$\varnothing$CoT} & \textbf{0S} & \textbf{$\varnothing$S} & \textbf{0CoT} & \textbf{$\varnothing$CoT}\\
\hline
PaLM 2 & 15.9\% (364) & \textcolor{myred}{3.28\% (75)} & 14.06\% (322) & \textcolor{myred}{7.69\% (176)} & 9.79\% (124) & \textcolor{myred}{0.87\% (11)} & 9.55\% (121) & \textcolor{myred}{3.95\% (50)}\\
PaLM 2 for Chat & 2.79\% (64) & 2.93\% (67) & 2.53\% (58) & 3.23\% (74) & 0.63\% (8) & \textcolor{myred}{0.47\% (6)} & 0.71\% (9) & \textcolor{myred}{0.63\% (8)}\\
Gemini Pro & 4.67\% (107) & \textcolor{myred}{2.4\% (55)} & 3.45\% (79) & \textcolor{myred}{3.28\% (75)} & 3.47\% (44) & \textcolor{myred}{2.45\% (31)} & 3.16\% (40) & \textcolor{myred}{1.26\% (16)}\\
Gemini Pro (Chat) & 4.19\% (96) & \textcolor{myred}{2.45\% (56)} & 3.49\% (80) & \textcolor{myred}{3.36\% (77)} & 3.55\% (45) & \textcolor{myred}{2.45\% (31)} & 3.16\% (40) & \textcolor{myred}{0.87\% (11)}\\
\hline
\end{tabular}

\begin{tabular}{rcc|cc|cc|cc}
\hline
\multirow{2}{*}{\textbf{Model}} & \multicolumn{4}{c|}{\textbf{RACE-m}} & \multicolumn{4}{c}{\textbf{RACE-h}}\\
& \textbf{0S} & \textbf{$\varnothing$S} & \textbf{0CoT} & \textbf{$\varnothing$CoT} & \textbf{0S} & \textbf{$\varnothing$S} & \textbf{0CoT} & \textbf{$\varnothing$CoT}\\
\hline
PaLM 2 & 6.82\% (98) & \textcolor{myred}{5.01\% (72)} & 8.43\% (121) & \textcolor{myred}{6.55\% (94)} & 15.21\% (532) & \textcolor{myred}{11.29\% (395)} & 15.78\% (552) & \textcolor{myred}{13.18\% (461)}\\
PaLM 2 for Chat & 3.2\% (46) & \textcolor{myred}{1.95\% (28)} & 3.41\% (49) & \textcolor{myred}{3.27\% (47)} & 3.69\% (129) & \textcolor{myred}{2.54\% (89)} & 3.6\% (126) & 3.77\% (132)\\
Gemini Pro & 5.43\% (78) & \textcolor{myred}{4.18\% (60)} & 6.34\% (91) & \textcolor{myred}{5.78\% (83)} & 6.2\% (217) & \textcolor{myred}{4.63\% (162)} & 6.38\% (223) & \textcolor{myred}{5.26\% (184)}\\
Gemini Pro (Chat) & 5.43\% (78) & \textcolor{myred}{4.11\% (59)} & 5.64\% (81) & \textcolor{myred}{4.87\% (70)} & 6.38\% (223) & \textcolor{myred}{4.75\% (166)} & 6.46\% (226) & \textcolor{myred}{5.37\% (188)}\\
\hline
\end{tabular}

\begin{tabular}{rcc|cc|cc|cc}
\hline
\multirow{2}{*}{\textbf{Model}} & \multicolumn{4}{c|}{\textbf{ANLI}} & \multicolumn{4}{c}{\textbf{TriviaQA}}\\
& \textbf{0S} & \textbf{$\varnothing$S} & \textbf{0CoT} & \textbf{$\varnothing$CoT} & \textbf{0S} & \textbf{$\varnothing$S} & \textbf{0CoT} & \textbf{$\varnothing$CoT}\\
\hline
PaLM 2 & 8.83\% (106) & \textcolor{myred}{3.92\% (47)} & 8.42\% (101) & 8.67\% (104) & 10.2\% (102) & \textcolor{myred}{2.8\% (28)} & 6.7\% (67) & \textcolor{myred}{5.4\% (54)}\\
PaLM 2 for Chat & 0.33\% (4) & 0.42\% (5) & 0.08\% (1) & 0.42\% (5) & 4.8\% (48) & \textcolor{myred}{4.3\% (43)} & 4.3\% (43) & 5.7\% (57)\\
Gemini Pro & 1.92\% (23) & \textcolor{myred}{0.5\% (6)} & 1.5\% (18) & \textcolor{myred}{1.42\% (17)} & 5.7\% (57) & \textcolor{myred}{2.3\% (23)} & 4.2\% (42) & \textcolor{myred}{1.7\% (17)}\\
Gemini Pro (Chat) & 2.33\% (28) & \textcolor{myred}{0.83\% (10)} & 1.83\% (22) & \textcolor{myred}{1.67\% (20)} & 5.8\% (58) & \textcolor{myred}{2.3\% (23)} & 3.9\% (39) & \textcolor{myred}{1.3\% (13)}\\
\hline
\end{tabular}

\caption{This table displays the ratio of cases where each model responds with an empty string, representing instances where a generated response or a prompt is blocked by safety mechanisms built into the model's pipelines. \textcolor{myred}{Red color} represents a case where prompting decreases the number of empty responses. 0S, $\varnothing$S, 0CoT, and $\varnothing$CoT denote zero-shot prompting, \ns\ prompting, zero-shot CoT prompting, and \ns\ CoT prompting, respectively.}
\label{tab:empty_responses}
\end{table*}

We observe interesting results where the utilization of \ns\ or \nscot\ prompting decreases the effectiveness of the built-in safety mechanisms in the majority of cases. As shown in \autoref{tab:empty_responses}, the \textcolor{myred}{red color} highlights the decrease in the numbers of empty responses across datasets when eliciting through \ns\ or \nscot\ prompting. We find that both prompting can decrease empty responses by 25.02\% on average -- 44.77\% and 5.26\% on average when using \ns\ and \nscot\ prompting, respectively. We posit that the prompting distracts the models and deviates the models from usual patterns. Thus, the prompting decreases the effectiveness of the built-in safety measures. The results show a possibility to utilize both PE techniques to circumvent the safety mechanisms built into the models in a similar spirit to jailbreaking through prompting \citep{Chao:2023,Shen:2023}, but in our case, the safety breach is potentially at the API pipeline level. We conjecture that measures deployed during training, such as safe instruction-tuning, may not generalize enough to safeguard against all cases, in particular, when the models get distracted with hallucination-elicited prompts.

Comparing \ns\ and \nscot\ prompting, we notice that \ns\ prompting is more effective in breaking built-in safety measures, which is likely attributed to the fact that \nscot\ prompting induces reasoning during decoding. Therefore, we posit the same reasons for the reduced effectiveness of \nscot\ prompting in general; reasoning reduces the chances of hallucination. In addition, we observe that PaLM 2 for Chat generates fewer empty responses than PaLM 2 in general. In contrast, Gemini Pro is more consistent across text and chat generation. This observation may result from the fact that PaLM 2 and PaLM 2 for Chat are two different models, while both Gemini Pro variants are based on the same model. Nevertheless, we cannot confirm this fact due to a lack of public report.

\section{Absolute Results}\label{sec:absolute_results}
This section provides absolute results for the main experiment and ablation studies. \autoref{tab:full_results} presents absolute results from the main experiment. Tables \ref{tab:full_scaling_results}, \ref{tab:full_cot_results}, \ref{tab:full_position_results}, and \ref{tab:full_components_results} provide absolute results for the scaling analysis, the reasoning variant, the position analysis, and the prompt components analysis, respectively.

\begin{table*}[tbp]
\centering
\begin{tabular}{rcc|cc|cc|cc}
\hline
\multirow{2}{*}{\textbf{Model}} & \multicolumn{2}{c|}{\textbf{AQuA}} & \multicolumn{2}{c|}{\textbf{GSM8K}} & \multicolumn{2}{c|}{\textbf{StrategyQA}} & \multicolumn{2}{c}{\textbf{WinoGrande}}\\
& \textbf{0S} & \textbf{$\varnothing$S} & \textbf{0S} & \textbf{$\varnothing$S} & \textbf{0S} & \textbf{$\varnothing$S} & \textbf{0S} & \textbf{$\varnothing$S}\\
\hline
PaLM 2 & 29.13 & 28.35 & 14.78 & \underline{16.45} & 59.83 & \underline{66.38} & 72.69 & \underline{80.03}\\
PaLM 2 for Chat & 14.96 & \underline{15.75} & 53.90 & \underline{55.12} & 57.73 & \underline{58.69} & 56.59 & \underline{60.54}\\
GPT-3.5 Turbo & 42.91 & \underline{57.48} & 54.89 & \underline{63.23} & 64.02 & \underline{66.03} & 59.98 & 58.88\\
GPT-4 Turbo & 75.98 & 75.59 & 74.30 & 73.16 & 74.85 & 61.83 & 73.48 & 55.80\\
Gemini Pro & 25.59 & \underline{35.43} & 51.55 & \underline{66.49} & 67.03 & 50.66 & 63.85 & 62.98\\
Gemini Pro (Chat) & 25.59 & \underline{37.01} & 52.39 & \underline{67.02} & 67.60 & 50.44 & 63.69 & 62.98\\
\hline
\end{tabular}
\begin{tabular}{rcc|cc|cc|cc}
\hline
\multirow{2}{*}{\textbf{Model}} & \multicolumn{2}{c|}{\textbf{RACE-m}} & \multicolumn{2}{c|}{\textbf{RACE-h}} & \multicolumn{2}{c|}{\textbf{ANLI}} & \multicolumn{2}{c}{\textbf{TriviaQA}}\\
& \textbf{0S} & \textbf{$\varnothing$S} & \textbf{0S} & \textbf{$\varnothing$S} & \textbf{0S} & \textbf{$\varnothing$S} & \textbf{0S} & \textbf{$\varnothing$S}\\
\hline
PaLM 2 & 82.66 & \underline{84.19} & 71.56 & \underline{74.16} & 49.17 & \underline{50.50} & 64.20 & \underline{68.70}\\
PaLM 2 for Chat & 73.54 & \underline{74.30} & 67.04 & \underline{67.50} & 42.67 & \underline{43.33} & 70.20 & 70.10\\
GPT-3.5 Turbo & 85.38 & 83.84 & 81.73 & 80.76 & 48.42 & 46.67 & 81.00 & \underline{82.00}\\
GPT-4 Turbo & 92.97 & \underline{93.25} & 88.59 & \underline{88.97} & 64.17 & 64.00 & 85.40 & 84.60\\
Gemini Pro & 83.01 & \underline{84.61} & 77.82 & \underline{79.47} & 50.58 & \underline{51.67} & 70.20 & 25.30\\
Gemini Pro (Chat) & 84.61 & \underline{85.24} & 79.10 & \underline{80.39} & 51.08 & \underline{51.92} & 70.50 & 25.40\\
\hline
\end{tabular}
\caption{Absolute results of the main experiment. \textbf{0S} denotes zero-shot prompting, \textbf{$\varnothing$S} denotes \ns\ prompting. \underline{Underline} signifies cases where \ns\ prompting results outperform its zero-shot prompting baseline.}
\label{tab:full_results}
\end{table*}

\begin{table*}[tbp]
\centering
\begin{tabular}{rcc|cc|cc|cc}
\hline
\multirow{2}{*}{\textbf{Model}} & \multicolumn{2}{c|}{\textbf{AQuA}} & \multicolumn{2}{c|}{\textbf{GSM8K}} & \multicolumn{2}{c|}{\textbf{StrategyQA}} & \multicolumn{2}{c}{\textbf{WinoGrande}}\\
& \textbf{0S} & \textbf{$\varnothing$S} & \textbf{0S} & \textbf{$\varnothing$S} & \textbf{0S} & \textbf{$\varnothing$S} & \textbf{0S} & \textbf{$\varnothing$S}\\
\hline
Llama 2 7B & 25.59 & 20.87 & 2.88 & \underline{3.18} & 50.79 & \underline{50.83} & 45.22 & 44.28\\
Llama 2 7B Chat & 52.75 & 49.65 & 21.53 & \underline{21.99} & 52.75 & 49.65 & 49.01 & 45.07\\
\hline
\end{tabular}
\begin{tabular}{rcc|cc|cc|cc}
\hline
\multirow{2}{*}{\textbf{Model}} & \multicolumn{2}{c|}{\textbf{RACE-m}} & \multicolumn{2}{c|}{\textbf{RACE-h}} & \multicolumn{2}{c|}{\textbf{ANLI}} & \multicolumn{2}{c}{\textbf{TriviaQA}}\\
& \textbf{0S} & \textbf{$\varnothing$S} & \textbf{0S} & \textbf{$\varnothing$S} & \textbf{0S} & \textbf{$\varnothing$S} & \textbf{0S} & \textbf{$\varnothing$S}\\
\hline
Llama 2 7B & 27.09 & \underline{30.50} & 23.90 & \underline{27.44} & 32.92 & \underline{34.17} & 64.80 & \underline{65.40}\\
Llama 2 7B Chat & 57.17 & 48.05 & 46.37 & 38.94 & 35.50 & 34.00 & 64.30 & 62.90\\
\hline
\end{tabular}
\caption{Absolute results of the scaling analysis experiment. \textbf{0S} denotes zero-shot prompting, \textbf{$\varnothing$S} denotes \ns\ prompting. \underline{Underline} signifies cases where \ns\ prompting results outperform its zero-shot prompting baseline.}
\label{tab:full_scaling_results}
\end{table*}

\begin{table*}[tbp]
\centering
\begin{tabular}{rcc|cc|cc|cc}
\hline
\multirow{2}{*}{\textbf{Model}} & \multicolumn{2}{c|}{\textbf{AQuA}} & \multicolumn{2}{c|}{\textbf{GSM8K}} & \multicolumn{2}{c|}{\textbf{StrategyQA}} & \multicolumn{2}{c}{\textbf{WinoGrande}}\\
& \textbf{0CoT} & \textbf{$\varnothing$CoT} & \textbf{0CoT} & \textbf{$\varnothing$CoT} & \textbf{0CoT} & \textbf{$\varnothing$CoT} & \textbf{0CoT} & \textbf{$\varnothing$CoT}\\
\hline
PaLM 2 & 35.43 & 16.14 & 60.20 & 43.52 & 62.71 & 59.34 & 63.93 & \underline{76.01}\\
PaLM 2 for Chat & 13.39 & 12.60 & 58.23 & 53.83 & 52.40 & \underline{60.13} & 59.83 & 57.14\\
GPT-3.5 Turbo & 57.48 & 55.51 & 66.41 & 63.76 & 66.11 & 57.38 & 51.14 & 45.54\\
GPT-4 Turbo & 75.59 & \underline{77.17} & 74.45 & 73.62 & 63.45 & 61.97 & 51.93 & \underline{64.80}\\
Gemini Pro & 46.46 & \underline{50.39} & 69.07 & 62.17 & 66.42 & 1.05 & 62.83 & 0.24\\
Gemini Pro (Chat) & 48.82 & \underline{52.76} & 70.89 & 62.62 & 66.42 & 1.05 & 62.43 & 0.24\\
\hline
\end{tabular}
\begin{tabular}{rcc|cc|cc|cc}
\hline
\multirow{2}{*}{\textbf{Model}} & \multicolumn{2}{c|}{\textbf{RACE-m}} & \multicolumn{2}{c|}{\textbf{RACE-h}} & \multicolumn{2}{c|}{\textbf{ANLI}} & \multicolumn{2}{c}{\textbf{TriviaQA}}\\
& \textbf{0CoT} & \textbf{$\varnothing$CoT} & \textbf{0CoT} & \textbf{$\varnothing$CoT} & \textbf{0CoT} & \textbf{$\varnothing$CoT} & \textbf{0CoT} & \textbf{$\varnothing$CoT}\\
\hline
PaLM 2 & 71.03 & \underline{82.52} & 60.12 & \underline{72.24} & 49.17 & 48.75 & 66.70 & 63.70\\
PaLM 2 for Chat & 71.59 & \underline{71.94} & 65.04 & \underline{65.55} & 43.50 & 42.42 & 70.60 & 69.10\\
GPT-3.5 Turbo & 83.36 & 80.57 & 77.67 & \underline{77.79} & 42.25 & 22.42 & 80.30 & \underline{80.90}\\
GPT-4 Turbo & 71.59 & 67.55 & 61.75 & 56.46 & 52.17 & 47.58 & 86.20 & 85.30\\
Gemini Pro & 83.15 & 80.64 & 78.47 & 77.64 & 48.33 & 44.92 & 61.60 & 0.90\\
Gemini Pro (Chat) & 84.05 & 82.10 & 78.36 & 77.79 & 48.92 & 45.00 & 62.00 & 0.90\\
\hline
\end{tabular}
\caption{Absolute results of the \nscot\ prompting experiment. \textbf{0S} denotes zero-shot prompting, \textbf{$\varnothing$S} denotes \ns\ prompting. \underline{Underline} signifies cases where \ns\ prompting results outperform its zero-shot prompting baseline.}
\label{tab:full_cot_results}
\end{table*}

\begin{table*}[tbp]
\centering\small
\begin{tabular}{rcccccccc}
\hline
\textbf{Prompting} & \textbf{AQuA} & \textbf{GSM8K} & \textbf{StrategyQA} & \textbf{WinoGrande} & \textbf{RACE-m} & \textbf{RACE-h} & \textbf{ANLI} & \textbf{TriviaQA}\\
\hline
0-shot & 42.91 & 54.89 & 64.02 & 59.98 & 85.38 & 81.73 & 48.42 & 81.00\\
\hline
\ns & \underline{57.48} & 63.23 & \underline{66.03} & \underline{58.88} & \underline{83.84} & \underline{80.76} & \underline{46.67} & \underline{82.00}\\
\ns-after & 55.91 & \underline{65.43} & 60.04 & 18.55 & 81.27 & 77.96 & 25.92 & 78.50\\
\hline
\end{tabular}
\caption{Absolute results of the effects of positions experiment. \textbf{0S} denotes zero-shot prompting, \textbf{$\varnothing$S} denotes \ns\ prompting. \underline{Underline} signifies cases where \ns\ prompting results outperform its zero-shot prompting baseline.}
\label{tab:full_position_results}
\end{table*}

\begin{table*}[tbp]
\centering\small
\begin{tabular}{rcccccccc}
\hline
\textbf{Prompting} & \textbf{AQuA} & \textbf{GSM8K} & \textbf{StrategyQA} & \textbf{WinoGrande} & \textbf{RACE-m} & \textbf{RACE-h} & \textbf{ANLI} & \textbf{TriviaQA}\\
\hline
0-shot & 42.91 & 54.89 & 64.02 & 59.98 & 85.38 & 81.73 & 48.42 & 81.00\\
\hline
\ns & 57.48 & 63.23 & 66.03 & 58.88 & 83.84 & 80.76 & 46.67 & \underline{82.00}\\
\hline
\ns-v1 & \underline{58.66} & \underline{64.14} & 65.76 & 57.62 & 83.57 & 80.30 & 45.33 & 81.50\\
\ns-v2 & 47.24 & 59.82 & \underline{66.94} & \underline{61.09} & \underline{84.40} & \underline{81.07} & \underline{48.67} & 80.70\\
\ns-v3 & 54.72 & 63.61 & 64.80 & 54.70 & 84.12 & 80.56 & 47.42 & 80.80\\
\hline
\end{tabular}
\caption{Absolute results of the effects of components experiment. \textbf{0S} denotes zero-shot prompting, \textbf{$\varnothing$S} denotes \ns\ prompting. \underline{Underline} signifies cases where \ns\ prompting results outperform its zero-shot prompting baseline.}
\label{tab:full_components_results}
\end{table*}

\section{Qualitative Examples of Null-Shot Prompting}\label{sec:examples}
In this section, we provide qualitative examples of generated responses from the datasets when utilizing \ns\ prompting. The LLM used to generate each response is denoted in the figure caption. Figures \ref{fig:aqua_example}, \ref{fig:gsm8k_example}, \ref{fig:strategyqa_example}, \ref{fig:winogrande_example}, \ref{fig:race_m_example}, \ref{fig:race_h_example}, \ref{fig:anli_example}, and \ref{fig:triviaqa_example} are examples of AQuA-RAT, GSM8K, StrategyQA, WinoGrande, RACE-m, RACE-h, ANLI, and TriviaQA, respectively.

\begin{figure*}[tbp]
\begin{tcolorbox}[colback=myblue!5!white,colframe=myblue]
  Look at examples in the ``Examples'' section and utilize examples and information from that section to perform the following task.\\
  Question: John likes to have lightly flavored tea every evening. In a 50\% strong milk tea, he replaces 15\% of it with milk twice. Then, he replaces 10 percent of the resultant solution with more milk.\\
  What is the final concentration of tea John drinks?\\
  Choices: A)15.38\%, B)42\%, C)39.86\%, D)22.35\%, E)32.51\%\\
  Answer:
  \tcblower
  To find the final concentration of tea John drinks, we can use the information from the ``Examples'' section.\\ 
  \\
  First, we need to find the concentration of tea after the first replacement of 15\% with milk. \\
  Initial concentration of tea = 50\%\\
  After first replacement = 50\% - (50\% * 15\%) = 50\% - 7.5\% = 42.5\%\\
  \\
  Then, we need to find the concentration of tea after the second replacement of 15\% with milk. \\
  Initial concentration of tea = 42.5\%\\
  After second replacement = 42.5\% - (42.5\% * 15\%) = 42.5\% - 6.375\% = 36.125\%\\
  \\
  Finally, we need to find the concentration of tea after the third replacement of 10\% with milk. \\
  Initial concentration of tea = 36.125\%\\
  After third replacement = 36.125\% - (36.125\% * 10\%) = 36.125\% - 3.6125\% = 32.5125\%\\
  \\
  So, the final concentration of tea John drinks is approximately 32.51\%. Therefore, the correct choice is \textcolor{mygreen}{\textbf{E) 32.51\%}}.
\end{tcolorbox}
\caption{An output generated using \ns\ prompting with GPT-3.5 Turbo for AQuA-RAT.}
\label{fig:aqua_example}
\end{figure*}

\begin{figure*}[tbp]
\begin{tcolorbox}[colback=myblue!5!white,colframe=myblue]
  Look at examples in the ``Examples'' section and utilize examples and information from that section to perform the following task.\\
  Question: On Mondays, Wednesdays, and Fridays, college student Kimo has three 1-hour  classes each day.  On Tuesdays and Thursdays, he has two 2-hour classes each day.  In one semester, there are 16 weeks of school.  In one semester, how many hours does Kimo spend attending classes?\\
  Answer:
  \tcblower
  To calculate the total hours Kimo spends attending classes in one semester, we can use the information provided in the examples section.\\
  \\
  On Mondays, Wednesdays, and Fridays:\\
  3 classes x 1 hour each = 3 hours per day\\
  3 hours x 3 days = 9 hours per week\\
  \\
  On Tuesdays and Thursdays:\\
  2 classes x 2 hours each = 4 hours per day\\
  4 hours x 2 days = 8 hours per week\\
  \\
  Total hours per week:\\
  9 hours (MWF) + 8 hours (TTh) = 17 hours per week\\
  \\
  Total hours for 16 weeks:\\
  17 hours/week x 16 weeks = 272 hours\\
  \\
  So, in one semester, Kimo spends \textcolor{mygreen}{\textbf{272 hours}} attending classes.
\end{tcolorbox}
\caption{An output generated using \ns\ prompting with GPT-3.5 Turbo for GSM8K.}
\label{fig:gsm8k_example}
\end{figure*}

\begin{figure*}[tbp]
\begin{tcolorbox}[colback=myblue!5!white,colframe=myblue]
  Look at examples in the ``Examples'' section and utilize examples and information from that section to perform the following task.\\
  Question: Can someone sell their time through the Toronto Star?\\
  Choices: A) True, B) False\\
  Answer:
  \tcblower
  \textcolor{mygreen}{\textbf{A) True}}\\
  \\
  Example from the ``Examples'' section: ``People can sell their time through the Toronto Star by offering services such as tutoring, dog walking, or house cleaning through classified ads or online platforms.''
\end{tcolorbox}
\caption{An output generated using \ns\ prompting with GPT-3.5 Turbo for StrategyQA.}
\label{fig:strategyqa_example}
\end{figure*}

\begin{figure*}[tbp]
\begin{tcolorbox}[colback=myblue!5!white,colframe=myblue]
  Look at examples in the ``Examples'' section and utilize examples and information from that section to perform the following task.\\
  Question: They were worried the wine would ruin the bed and the blanket, but the \_ was't ruined.\\
  Choices: 1) blanket, 2) bed\\
  Answer:
  \tcblower
  \textcolor{mygreen}{\textbf{2) bed}}\\
  \\
  Example from the ``Examples'' section: ``I spilled wine on the bed and the blanket, but the bed wasn't ruined.''
\end{tcolorbox}
\caption{An output generated using \ns\ prompting with GPT-3.5 Turbo for WinoGrande.}
\label{fig:winogrande_example}
\end{figure*}

\begin{figure*}[tbp]
\begin{tcolorbox}[colback=myblue!5!white,colframe=myblue]
  Look at examples in the ``Examples'' section and utilize examples and information from that section to perform the following task.\\
  Article: A boy and his father were walking in the mountains. Suddenly the boy fell,hurt himself,and cried,``AAAhhhhhhhhh!!!'' To his surprise,he heard the voice repeating ,somewhere in the mountain,``AAAhhhhhhhh!!!'' Then the boy shouted,``Who are you?'' He received the answer,``Who are you?''He got angry at the answer,so he shouted,  ``Foolish!''He received the answer,``Foolish!''\\
  He looked at his father and asked,``What'sgoing on?''The father smiled and said,``My son,listen,''And then he shouted to the mountain,``Ilove you!''The voice answed,``Ilove you!'' Again the man cried,``You are the best!'' The voice answed,``You are the best!''\\
  The boy was surprised,but did not understand. Then the father explained ,``People call this 'ECSO',but really this is Life.It gives you back everything you say or do. Our life is just a reflection of what we have done. If you  want more love in the world,have more love in your heart. If you want to be successful,work hard. This can be used in everything in life. Life will give you back everything you have given to it.\\
  Question: When the boy heard the voice repeating,he thought  \_  .\\
  Choices: A) it was foolish to hear others' voice, B) someone else in the mountain liked his voice, C) there were many other people in the mountain, D) he was laughed at by someone else in the mountain\\
  Answer:
  \tcblower
  \textcolor{mygreen}{\textbf{D) he was laughed at by someone else in the mountain}}
\end{tcolorbox}
\caption{An output generated using \ns\ prompting with GPT-4 Turbo for RACE-m.}
\label{fig:race_m_example}
\end{figure*}

\begin{figure*}[tbp]
\begin{tcolorbox}[colback=myblue!5!white,colframe=myblue]
  Look at examples in the ``Examples'' section and utilize examples and information from that section to perform the following task.\\
  Article: Japan' s nuclear crisis is also causing concerns in China. Worried shoppers rushed to stores and supermarkets in the country for salt, with the belief that it might protect them from potential nuclear radiation spreading from Japan.\\
  Many rushed to buy to store as much iodized salt as possible because of the rumor that iodized salt could help prevent radiation poisoning spreading from Japan' s Fukushima Daiichi power plant. Consumers also worried that the nuclear plumes might spread  to China by air and sea, polluting food sources, including salt taken from the sea.\\
  The wave of panic buying spread quickly across the country, driving up salt prices by five to ten fold in some cities.\\
  Even regions rich in salt production, like Jiangxi Province,faced sudden shortages.\\
  The government took measures to guarantee enough supply and to stabilize the market. It also dismissed the nuclear rumors, saying the nuclear fallout was unlikely to reach the country, and that salt did not help to prevent radiation poisoning.\\
  Local authorities have stepped in to stabilize the market, too. As well, state-owned salt companies have been urged to increase their supply.\\
  Facing expanding market demand, distribution centers have taken extra measures to ensure they are well-prepared.\\
  Wang Yun, the General Manager of Beijing Salt Industry Corporation, said, ``Our storage reserves can guarantee a two-month supply for the Beijing market.''\\
  Question: People made panic purchases of salt for the reasons EXCEPT  \_  .\\
  Choices: A) salt might protect them from potential nuclear radiation, B) nuclear radiation might pollute food sources, C) nuclear radiation might pollute salt taken from the sea, D) there would be a shortage of salt on the market\\
  Answer:
  \tcblower
  \textcolor{mygreen}{\textbf{D) there would be a shortage of salt on the market}}
\end{tcolorbox}
\caption{An output generated using \ns\ prompting with GPT-4 Turbo for RACE-h.}
\label{fig:race_h_example}
\end{figure*}

\begin{figure*}[tbp]
\begin{tcolorbox}[colback=myblue!5!white,colframe=myblue]
  Look at examples in the ``Examples'' section and utilize examples and information from that section to perform the following task.\\
  Context: How to handle smart people<br>Accept that you cannot always be the smartest person in the room. There will probably be people who are able to comprehend certain things easier than you. Everyone is talented in their own way and are good and bad at different things.\\
  Hypothesis: ANgelina never takes the kids to school\\
  Question: Does the hypothesis imply `entailment,' `neutral,' or `contradiction' in the given context?\\
  Choices: A) entailment, B) neutral, C) contradiction\\
  Answer:
  \tcblower
  The correct answer is: **\textcolor{mygreen}{\textbf{B) neutral}}**\\
  \\
  The hypothesis ``Angelina never takes the kids to school'' is neutral in the given context. This is because the hypothesis does not provide any information about whether or not Angelina is smart. It is possible that Angelina is smart and chooses not to take the kids to school, or that Angelina is not smart and chooses not to take the kids to school. The hypothesis does not provide enough information to determine whether or not there is an entailment, contradiction, or neutral relationship between the hypothesis and the context.\\
  \\
  Here are some examples of entailment, contradiction, and neutral relationships:\\
  \\
  * **Entailment:** If it is raining, then the ground is wet. This is an example of entailment because the conclusion (the ground is wet) necessarily follows from the hypothesis (it is raining).\\
  * **Contradiction:** If it is raining, then the ground is dry. This is an example of contradiction because the conclusion (the ground is dry) contradicts the hypothesis (it is raining).\\
  * **Neutral:** If it is raining, then the ground is either wet or dry. This is an example of a neutral relationship because the conclusion (the ground is either wet or dry) does not necessarily follow from the hypothesis (it is raining).
\end{tcolorbox}
\caption{An output generated using \ns\ prompting with PaLM 2 for Chat for ANLI.}
\label{fig:anli_example}
\end{figure*}

\begin{figure*}[tbp]
\begin{tcolorbox}[colback=myblue!5!white,colframe=myblue]
  Look at examples in the ``Examples'' section and utilize examples and information from that section to perform the following task.\\ 
  Question: Who was named the best player at Euro 2012 by UEFA ?\\
  Answer:
  \tcblower
    The best player at Euro 2012 was named \textcolor{mygreen}{\textbf{Andres Iniesta}} by UEFA. This can be seen in the example from the ``Examples'' section: ``Andres Iniesta was named the best player at Euro 2012 by UEFA.''
\end{tcolorbox}
\caption{An output generated using \ns\ prompting with GPT-3.5 Turbo for TriviaQA.}
\label{fig:triviaqa_example}
\end{figure*}

\section{Raw Data and Source Code}

Raw data and source code will be made public upon acceptance.

\section{Declaration of AI Assistance}

We utilized ChatGPT only for grammatical checking and LaTeX support of the content presented in this study but did not use it for the initial draft of this study. GitHub Copilot was utilized for trivial and boilerplate code completion during data generation and data analysis. We declare that all content presented and code utilized in this study has been reviewed and edited by the authors.

\end{appendices}
\end{document}